\documentclass[aps,pre,twocolumn,letterpaper,floatfix]{revtex4-1}
\usepackage{algorithm}
\usepackage{algpseudocode}
\usepackage{graphicx}
\usepackage{amssymb}
\usepackage{amsmath}
\usepackage{mathrsfs}
\usepackage{placeins}
\usepackage{caption3}
\usepackage{wrapfig}
\usepackage{caption} 
\usepackage{subcaption}
\usepackage{xcolor}
\usepackage{contour}
\usepackage{color}

\usepackage[bookmarksnumbered=true,breaklinks=true,colorlinks,citecolor=blue,linkcolor=blue]{hyperref}
\newcommand{\reals}{\mathbb{R}} %this requires amssymb if not using txfonts

\newcommand{\trans}{^\mathrm{T}}
\newlength{\figwidth}
\newlength{\SCwidth}
\def\XXint#1#2#3{{\setbox0=\hbox{$#1{#2#3}{\int}$}
		\vcenter{\hbox{$#2#3$}}\kern-.5\wd0}}

\usepackage[normalem]{ulem}
\newcommand{\MDGrevise}[1]{\textcolor{black}{#1}}
\definecolor{ao(english)}{rgb}{0.0, 0.5, 0.0}

\newcommand{\ALrevise}[1]{\textcolor{black}{#1}}

\newcommand{\IM}{\mathcal{M}}
\newcommand{\chitilde}{\check{\chi}}
\begin{document}

%\preprint{}

%\title{Predicting Chaotic Dynamics of the Kuramoto-Sivashinsky Equation on an Inertial Manifold via Deep Learning}
\title{Deep learning to discover and predict dynamics on an inertial manifold} %MDG
%\title{Deep learning to discover an inertial manifold coordinate system and predict dynamics of the Kuramoto-Sivashinsky equation} 


\author{Alec J. Linot}
%\thanks{}
\affiliation{Department of Chemical and Biological Engineering, University of Wisconsin-Madison, Madison WI 53706, USA}

\author{Michael D. Graham}
%\thanks{}
\email{mdgraham@wisc.edu}
\affiliation{Department of Chemical and Biological Engineering, University of Wisconsin-Madison, Madison WI 53706, USA}
\date{\today}
\begin{abstract}
A data-driven framework is developed to represent chaotic dynamics on an inertial manifold (IM), and applied to solutions of the Kuramoto-Sivashinsky equation. A hybrid method combining linear and nonlinear (neural-network) dimension reduction transforms between coordinates in the full state space and on the IM.  Additional neural networks predict time-evolution on the IM. The formalism accounts for translation invariance and energy conservation, and substantially outperforms linear dimension reduction, reproducing very well key dynamic and statistical features of the attractor.  
\end{abstract}
%\keywords{} 
\maketitle
%2
% Describe work done towards creating physics informed NN
%\section{Introduction} \label{Introduction}\MDG{no separate headings in PRL -- they take up too much space}
% Discuss NN in general
%\MDG{most of this is irrelevant to the current work.  Start by talking about how the dynamics of dissipative partial differential equations like the NSE live on finite-dimensional inertial manifolds. What I like to do for the first paragraph of a paper is to briefly introduce in general terms the problem we're trying to tackle, then briefly describe the approach we take in the paper to do so.}
\section{Introduction}
Partial differential equations are formally infinite-dimensional, but the presence of dissipation (through viscosity or diffusion, for example) leads to the expectation that the long-time dynamics collapse onto a finite-dimensional invariant manifold \cite{Hopf1948a}. Specifically, for some systems, including the Kuramoto-Sivashinsky equation (KSE) that we consider here, it can be proven that all initial conditions exponentially approach an \emph{inertial manifold} $\mathcal{M}$ of finite dimension $d_\IM$ \cite{Foias1988a}, on which the long time dynamics evolve. 
%\ALrevise{Proving the existence of an IM remains an open problem for the NSE, but, empirically, simulations of the NSE tend to decay to an attractive region with a finite fractal dimension \cite{Keefe1992}. Also, recent work suggests NSE in two dimensions has an IM, along with the hyperviscous NSE in three dimensions \cite{Li2019}.}
%\ALrevise{For PDEs that satisfy the spectral gap condition an inertial manifold will exist \cite{Zelik2013}.}\AL{\textbf{The specific Theorem is 2.1 in Zelik. Should I cite that?}CITE THEOREM FOR THIS.}
%\ALrevise{Given any initial condition $u_0\in H$, where $H$ is the original state space, for a system with an inertial manifold the state will evolve to $u\in \IM$ exponentially. 
For states on the IM, $v\in \IM$, one can in principle find a coordinate transformation $h=\chi(v)$ to coordinates $h$ on the inertial manifold, a change of coordinates $v=\chitilde(h)$ back to the full space, and a dynamical system $h(t+\tau)=F(h(t))$  on $\IM$. (Alternately, one could represent the dynamical system in differential form $dh/dt=G(h)$.) This dynamical system is an exact reduced-order model (ROM). Such a model can be practically useful, for computationally  efficient simulations of a complex process, and may also be fundamentally important, since the coordinates $h$ represent the key dynamical variables for the phenomenon of interest. 
%\ALrevise{However, interpreting data-driven representations remains an open problem that we will not explore here.}\MDG{without context, this is out of place -- I'm not sure this is how/where we should address the interpretability question}

In the present work, we use ``data'' in the form of chaotic solutions to the KSE with periodic boundary conditions, to find neural-network (NN) representations of the functions $\chi, \chitilde$, and $F$. Many prior studies of inertial manifolds, and approximations thereof, take the inertial manifold to be the graph of a function $\Phi$ such that $\mathcal{M}:=\{v_++\Phi(v_+)\}$, where $v_+=Pv$ is a projection onto the $d_\IM$ leading eigenfunctions of the linear operator for the PDE \cite{Zelik2013,Foias1988,Graham:1993wv}. The present work is not subject to this restriction. Furthermore, our formalism explicitly accounts for the important physical features of translation invariance and energy conservation found in this system, and reproduces, very well, with a minimal number of degrees of freedom, key dynamic and statistical features of the attractor. 

A standard machine learning method for nonlinear dimension reduction is the undercomplete autoencoder \cite{Hinton:2006bg,Goodfellow2016}. This is a pair of neural networks, one mapping from a high-dimensional space to a low-dimensional one, and the second doing the reverse. The networks take data $u$ as input, compute an output $\tilde{u}$ and are trained to minimize a loss function $L=||u-\tilde{u}||^2$ summed over a batch of data vectors $u$. %\MDG{does the loss contain the sum?} \AL{Yes.}
Autoencoders have been used for nonlinear dimension reduction in many applications \citep{Hinton:2006bg}, including turbulent flow fields \cite{Omata:2019ho,Milano2002}. For dynamical systems, autoencoders have been used to explicitly yield coordinate transformations on which the dynamics are linear \citep{Otto2019,Lusch:2018ex} (e.g.~to determine eigenmodes of the Koopman operator) or have a sparse representation \citep{Champion:2019vx}. Gonzalez et al.~\citep{Gonzalez:2018tj} have combined autoencoders with nonlinear time-evolution models for reconstruction of dynamics of isotropic turbulence and lid-driven cavity flow. Lee and Carlberg \citep{Lee2019} give an overview of nonlinear model reduction with NNs and application to transient dynamics of Burgers equation. The physical interpretation of NN  representations of physical phenomena has been explored in Iten et al. \cite{Iten2020}.

Other studies have focused on developing NNs for evolving equations with chaotic dynamics without nonlinear dimension reduction. \ALrevise{An early example of this is Gonz\'alez-Garc\'ia et al. \citep{Gonzalez-Garcia1998}, who used a NN to predict the right hand side of the discretized KSE. Specificallly, they input the full state of the KSE and its derivatives into a NN to predict the parameters of a Runge-Kutta method for time integration. More recently larger NN have been used for prediction.} For example, Pathak et al. \citep{Pathak:2018dg} showed that a reservoir network trained with time-evolution ``data" from the KSE was capable of making excellent predictions of future time-evolution. No explicit model reduction was performed. In \citep{Lui2019}, proper orthogonal decomposition (POD) and a spectral version (SPOD), both linear dimension reduction techniques, were used to reduce the dimension of fluid flow data that were then used to train a NN for time-evolution. 
%\ALrevise{This approach captures the main behavior, but resulted in the loss of high wavenumber modes because keeping the additional information needed to estimate these higher modes was computationally prohibitive.} \MDG{what resulted in the loss of these modes? Elaborate} This technique of linear dimension reduction has also been used for evolving other chaotic equations like the KSE. 
Vlachas et al.~\citep{Vlachas:2018gc} combined various linear dimension reduction approaches with a long-short term memory NN and mean stochastic modeling to keep trajectories on the attractor. Similarly, in \citep{Wan:2018ga} a long-short term memory NN was used in a nonlinear Galerkin approach to estimate the nonlinearity, a task often achieved by assuming lower modes evolve slowly, and iteratively solving for the higher modes \citep{Titi1990,Foias1988b,Jolly1990,Graham:1993wv,Matthies2003}. %\ALrevise{The latter approach, often called an approximate inertial manifold, has been applied to both the KSE \citep{Jolly1990} and the Navier-Stokes Equation \citep{Titi1990}.}
%\MDG{awkward transition here}%\ALrevise{Specifically for strictly data-driven methods of predicting the KSE, Pathak et al. \citep{Pathak:2018dg} showed excellent predictive capabilities using reservoir learning.} 
%Pathak et al. \citep{Pathak:2018dg} showed reservoir learning to be a highly effective method of predicting dynamics of the KSE. 

Although methods exist for modeling dynamics on an IM, estimating $d_\IM$ remains a difficult problem. In \cite{Kuptsov2019}, an autoencoder is used to estimate $d_\IM$ from data for the dynamics of the complex Ginzburg-Landau (CGL) equation; however the dynamics on $\IM$ are not modeled. 
%\sout{\ALrevise{Likewise, \citep{Maus2011} estimates $d_\IM$ for low dimensional chaotic maps \MDG{this needs a rewrite. Inertial manifolds are a feature of PDEs, not low-dimensional maps.  Also what is meant by ``low" here?} by inputting time delayed coordinates into a NN that reduces dimension and predicts the single step prediction. This work does not analyze predictive capabilities past single step predictions.}} 
A dynamical approach to determining $d_\IM$ was taken in \citep{Yang2009}, where the covariant Lyapunov vectors of trajectories the KSE and the CGL were found to decompose into ``physical" and ``isolated" modes. Physical modes are entangled, in the sense that tangencies between them result in perturbations of a single mode effecting other modes, whereas isolated modes lack tangencies with physical or isolated modes. This suggests the number of physical modes corresponds to $d_\IM$. Expanding on that work, Ding et al. \citep{Ding2016} found the dimension for the KSE in similar ways using Floquet vectors from an ensemble of unstable periodic orbits that are close to the chaotic attractor.
%\MDG{4/8: The following is incorrect. Chaotic orbits are very close to infinitely many unstable orbits of arbitrarily large period -- they do not hop between these orbits} This work supports the view that chaotic dynamics can be described as trajectories hopping between unstable periodic orbits.
The present work combines data-driven dimension reduction and time evolution using an efficient autoencoder structure that incorporates translation symmetry and energy conservation.

\section{Formulation} 
Our testbed for this approach is the KSE,
\begin{equation} \label{eq:KSE}
	\partial_t v=-v\partial_x v-\partial_{xx} v-\partial_{xxxx} v,
\end{equation}
with periodic boundary conditions in the domain $x\in[0,L]$. We select $L=22$, $44$, and $66$ because these domain sizes yield increasingly chaotic dynamics, and $d_\IM$ is known at $L=22$ \cite{Ding2016}.  Solutions to this equation are only unique to within a translation that we will represent with a phase variable $\phi\in \mathbb{R}$. This equation has an energy conservation principle: when time-averaged, the energy production rate $\mathcal{P}=\left<\partial_x v\partial_x v\right>$  balances the dissipation rate  $\varepsilon=\left< \partial_{xx} v\partial_{xx} v\right>$. Here $\langle\cdot\rangle$ represents averaging over $x$. 
% Here negative diffusion $\partial_{xx} u$ inputs energy into the system, hyperdiffusion $\partial_{xxxx} u$ dissipates energy, and convection enters in by $u\partial_x u$. \ALrevise{These conclusions appear from multiplying Equation \ref{eq:KSE} by $u$ resulting in an energy balance where the energy at a given time is given by $E=P-D$; here the spatially average production and dissipation are given by $P=\left<\partial_x u\partial_x u\right>$ and  $D=\left< \partial_{xx} u\partial_{xx} u\right>$. 
These properties are incorporated into the dimension reduction formulation as detailed below. Trajectories of \eqref{eq:KSE} were generated using a Fourier spectral method in space and a fourth-order time integration scheme \cite{Kassam2005} with the code available from Cvitanovi\'c et al. \cite{ChaosBook}.
%The solution \ALrevise{$v$ is represented on a uniformly spaced mesh of by $u\in\mathbb{R}^d$, where} the dimension $d$ of the full state space is $d=64$.
The solution $v(x)$ is represented on a uniformly-spaced mesh of $d=64$ points; we denote the solution on this mesh as $u\in\reals^d$, so $d$ is the dimension of the full state space in the present system.

\begin{figure} 
	\includegraphics[trim=0 0 0 0,width=8.6 cm,clip]{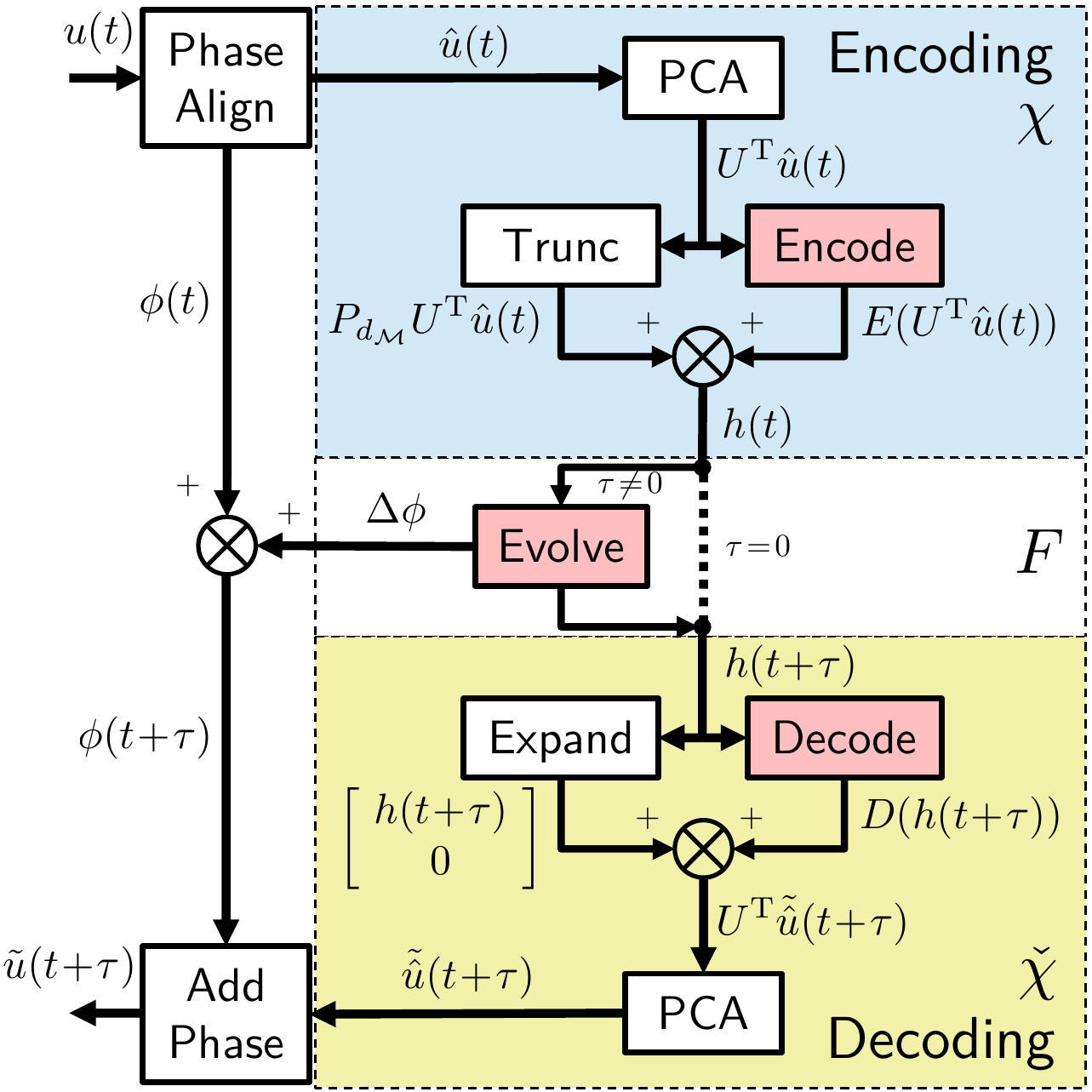}
	\captionsetup{justification=raggedright}
	\caption{Block diagram for the hybrid autoencoder and time-evolution scheme. NNs are pink (light gray).}
	\label{fig:Framework}
	\vspace{-5mm}
\end{figure} 

\section{Methodology and Results}
Figure \ref{fig:Framework} illustrates our framework for finding the inertial manifold and the dynamical system on it. The first step of the process exploits translation invariance: the solution $u$ at every time instant is transformed into a pattern $\hat{u}\in\mathbb{R}^d$ and a phase $\phi$ using an approach called the ``method of slices" \citep{Budanur2015b,Budanur2015}. Factoring out the phase leads to a more compact representation of the data by eliminating the need of training redundant weights for translated signals. (E.g., the representation will not need to separately represent $\sin 2\pi x/L$ and all of its translations.) Furthermore, such ``symmetry reduction" methods have been found to help elucidate the state space structure in fluid mechanics problems such as pipe flow \citep{Willis2013}.
%  We now describe this framework step by step. \MDG{done to here with methodology 12/3}
% First, an initial condition ($u_t \in \mathbb{R}^d$) is mapped to a translationally invariant space ($\hat{u}_t\in\mathbb{R}^{d}$), with phase $\phi_t\in\mathbb{R}$, using the method-of-slices (MOS) \citep{Budanur2015b,Budanur2015}. This step treats signals translated to different points in the domain equally, which prevents the need of training redundant weights for translated signals. Following the MOS, a NN reduces dimensions via the encoding portion of an autoencoder onto the inertial manifold ($h_t\in\mathbb{R}^{d_h}$). Then, in this space, the state ($h_{t+\tau}$) and phase ($\phi_{t+\tau}$) are evolved forward in time by separately trained NNs. After evolving the data on the inertial manifold a decoder maps data back to the invariant space ($\hat{u}_{t+\tau}$), and that snapshot is rotated to the new phase ($u_{t+\tau}$). Structures for the various NN functions appear in table \ref{Table}.

The method of slices involves taking the discrete Fourier transform $\mathcal{F}$ of the data in $x$ to yield $a(t)=\mathcal{F}\left\{u(t)\right\}$. With the data in Fourier space, the phase of the first Fourier mode $a_1(t)$ is found using 
%\begin{equation} \label{eq:phase}
$
\phi(t)=\text{atan2}(\text{Im}(a_{1}(t)),\text{Re}(a_{1}(t))).
$
%\end{equation}
Now we can construct a phase-aligned solution $\hat{u}(t)$ so that its first Fourier mode is a pure cosine:
% \begin{equation} \label{eq:shift}
$
\hat{u}(t)=\mathcal{F}^{-1}\left\{a(t)e^{-i k \phi(t)}\right\}.
$
% \end{equation}
Storage of $\phi$ for each time instant allows conversion of $\hat{u}$ back to $u$ (i.e.~$u(t)$ contains both $\phi(t)$ and $\hat{u}(t)$). Times when $a_1(t)$ approaches zero require special treatment, as was recognized by Budanur et al. \citep{Budanur2015b}. We use the solution they proposed, which is to stretch time according to $\Delta \hat{t}=\Delta t / |a_{1}(t)|$. The rescaled time $\hat{t}$ is called ``in-slice" time. Given any $u(t)$ we can always find $\hat{u}(t)$, $\phi(t)$ and $a_1(t)$, so it is always possible to move back and forth between the original and phase-aligned solutions and between real and in-slice time.
The data used for training the NNs were $\sim4-5\cdot 10^5$  solutions $u$ separated by $\Delta \hat{t}=0.2$ ``in-slice" time units, which corresponds to $\Delta t\approx0.023$ time units on average for $L=22$. Data was gathered after the dynamics had settled onto the attractor.

Given the phase-aligned pattern data $\hat{u}$, the first machine learning task is to find the manifold $\IM$, of dimension $d_\IM$, on which this data lives, or equivalently the coordinate tranformations $h=\chi(\hat{u})$ and $\hat{u}=\chitilde(h)$. Because the phase of any given data vector $u$ is arbitrary, the phase information is not needed for this step. The coordinates $h(t)$ and the phase $\phi(t)$ completely describe the state of the system so the dimension  of the attractor in the unreduced state space will be $d_\IM+1$. (For further discussion of invariant manifolds in translation-symmetric systems, see Ref. \cite{Budanur2015}.) Indeed, phase alignment allows more efficient representation of the data, because phase information need not be encoded -- it is captured separately as noted above.  

To find $\chi$ and $\chitilde$, we use a variant of a standard undercomplete autoencoder, shown as the ``$\tau=0$" branch of architecture shown in Figure \ref{fig:Framework}. This variant uses a NN to represent the \emph{difference} between the data and its projection onto the basis arising from principal components analysis (PCA) of the data set \cite{Strang2019}. PCA is widely used for linear dimension reduction because  it yields the projection of dimension $d_h$ that minimizes the mean squared deviation from the original data. Let $U$ be a square orthogonal matrix whose columns are the PCA basis vectors, and $P_{d_h}U\trans$ and $P_{d-d_h}U\trans$ the projection onto the first $d_h$ and last $d-d_h$ such vectors, respectively. The encoding step learns the function $E(U\trans\hat{u}(t))$ such that 
\begin{equation} \label{eq:hidden}
	E(U\trans\hat{u}(t))=h(t)-P_{d_h}U\trans\hat{u}(t).
\end{equation}
This structure is shown inside the blue (upper) box in Figure \ref{fig:Framework}. It must be emphasized that there is no approximation in choosing this representation. Furthermore, $U$ need not come from PCA; for example, $U$ could be the discrete Fourier transform operator or simply the identity, the latter corresponding to using the solution values on the mesh points.
%Furthermore,  $U$ need not come from PCA -- any other set of basis vectors, \ALrevise{the identity or} Fourier modes for example, could also be used.

%The goal of the autoencoder in this work is to map data onto the inertial manifold $h_t \in \mathbb{R}^{d_h}$ where the long-time dynamics exist, which is accomplished by adding a linear dimension reduction to a nonlinear dimension reduction 
%\begin{equation} \label{eq:hidden}
%h_t=\hat{u}_t^{\prime}+E(U\hat{u}_t).
%\end{equation}
%Although an autoencoder with appropriate width, depth, and activation ought to recreate any function selecting these parameters, properly initializing weights, and the effectiveness of the optimization method all hinder successfully approximating functions in practice. To overcome this issue we designed \ALrevise{a hybrid autoencoder (HNN) that modifies some linear map. In this case we chose to modify PCA because PCA provides the optimal linear basis for minimizing the reconstruction error.}\sout{an autoencoder with difference from PCA (PCANN) structure to combined the benefits of linear and nonlinear dimensionality reduction.} Figure \ref{fig:Framework} shows \ALrevise{these methods are combined by enriching the first $d_h$ PCA coefficients $u_t'$ with values from an encoder $E(U\hat{u}_t)$, where $U$ is the PCA basis calculated by taking the singular value decomposition of a snapshot matrix.} 
\begin{table}
	\captionsetup{justification=raggedright}
	\caption{Architectures of the NNs. ``Shape" indicates the dimension of each layer, and ``activation" the corresponding activation functions (S is the sigmoid activation) \cite{Goodfellow2016}. Decoder1 refers to domains $L=22$ and $44$. Decoder2 refers to $L=66$.}
	\resizebox{.45\textwidth}{!}{%
		\begin{tabular}{l*{6}{c}r}
			& Function & Shape & Activation \\
			\hline
			Encoder		& $E$ & $d:500:d_h$ & S:tanh  \\
			Decoder1		& $D$ & $d_h:500:d$  & S:linear  \\
			Decoder2		& $D$ & $d_h:500:500:d$  & S:linear  \\	
			Evolution	& $F_h$ & $d_h:200:200:d_h$ & S:S:linear  \\
			Evolution		 & $F_\phi$ & $d_h:500:50:500:1$ & S:S:S:linear  \\
			\label{Table}
	\end{tabular}}
	\vspace{-5mm}
\end{table}

The decoding step takes the data $h(t)\in\mathbb{R}^{d_h}$ in the inertial manifold coordinates and transforms it back to the full space, as shown in the yellow (lower) box in Figure \ref{fig:Framework}. 
%This is accomplished by adding the output of a decoding NN ($D(h_t)$) to the nonlinear projection ($[h_t,0]\trans$) and then changing the basis back. The output is then given by {\MDGrevise{Why? 
Again one can think of learning a difference: the decoder learns a function $D(h(t))$ such that
\begin{equation} \label{eq:Autoencoder}
	%\tilde{\hat{u}}_t = \chitilde(h_t)=U\trans\left( \left[\begin{array}{c}{h_t} \\ {0}\end{array}\right]+D(h_t)\right).
	D(h(t))=U\trans\tilde{\hat{u}}(t) -\begin{bmatrix}
		{h(t)} \\ 0
	\end{bmatrix}.
\end{equation}
%In this form it is clear that our method is the same as the original IM formulation\MDG{BUT IT'S NOT! The original formulation requires the nonlinear manifold to be parametrizable by a linear one -- you're undercutting a key point with this phrasing-- rewrite}, where instead of a linear projection $u_+$ there is the nonlinear projection $h$. Additionally, it is worth noting that i
Taking $E(U\trans\hat{u}(t))=0$ recovers the original IM formulation, but precludes the representation of curved manifolds that do not have a one-to-one mapping from a linear projection. An example of such a manifold is the Archimedean spiral, whose Cartesian representation is $(x,y)=(\phi\cos\phi,\phi\sin\phi$). The autoencoder architecture used here is able to represent this manifold with $d_h=1$.
% , shown in Figure \ref{fig:SwissRoll}. Our framework recreates data on the swiss roll overcoming the drawback of the linear projection, which comes from the fact that $D$ is a function of $h_t$, not just $P_{d_\IM}U\hat{u}_t$.

Finally, inserting Eq.~\ref{eq:hidden} into Eq.~\ref{eq:Autoencoder}, solving for $\tilde{\hat{u}}_t$, and noting that this can be written $\tilde{\hat{u}}(t) =U \left[P_{d_h}U\trans\hat{u}(t), {P_{d-d_h}U\trans\hat{u}(t)}\right]\trans$,
%By expanding $h_t$ and splitting $D$ in Equation \ref{eq:Autoencoder} we see that
%\begin{equation}
%\tilde{\hat{u}}_t =U \left[\begin{array}{c}{P_{d_\IM}U\trans\hat{u}_t+E(U\trans\hat{u}_t)+D_{d_\IM}\left(h_t\right)} \\ {D_{d-d_\IM}\left(h_t\right)}\end{array}\right].
%\end{equation}
%The exact solution can be written as 
%\begin{equation}
%\tilde{\hat{u}}_t =U \left[\begin{array}{c}{P_{d_\IM}U\trans\hat{u}_t} \\ {P_{d-d_\IM}U\trans\hat{u}_t}\end{array}\right],
%\end{equation}
shows that the exact solution satisfies $E(U\trans\hat{u}(t))+D_{d_h}(h(t))=0$, where $D_{d_h}$ contains the first $d_h$ components of $D$.
%and $D_{d-d_\IM}\left(h_t\right)=P_{d-d_\IM}U\trans\hat{u}_t$. 
%Unfortunately, explicitly enforcing $D_{d_\IM}(h_t)=E(U\trans\hat{u}_t)$ means the decoder requires the full state space vector, which defeats the purpose of dimension reduction.
This constraint can be satisfied approximately by adding a penalty term to the autoencoder loss function so it becomes
\begin{equation} \label{eq:Loss}
	L=||\hat{u}(t)-\tilde{\hat{u}}(t)||^2+\alpha ||E(\hat{u}(t))+D_{d_h}\left(h(t)\right)||^2.
\end{equation}
%Including this penalty improved performance. 
%In general, this penalty need not be added to the loss, but the PCA coefficients are organized according to importance for reconstruction, so accurately estimating the leading coefficients improved results. 
With this structure, we can in principle achieve an exact representation (within the approximation error of the functions $E$ and $D$) of data on a manifold of dimension $d_\IM$ for all $d_h\geq d_\IM$. \ALrevise{In general, the functions $E$ and $D$, or more generally $\chi$ and $\breve{\chi}$, need not come from NNs. Other approaches to nonlinear dimension reduction and function approximation (e.g.~tSNE, diffusion maps, kernel regression \citep{vanDerMaaten2008,vanDerMaaten2009}) might be useful as well. The overall structure of our approach would be the same.}

\begin{figure} 
	\includegraphics[trim=0 0 0 0,width=8.6 cm,clip]{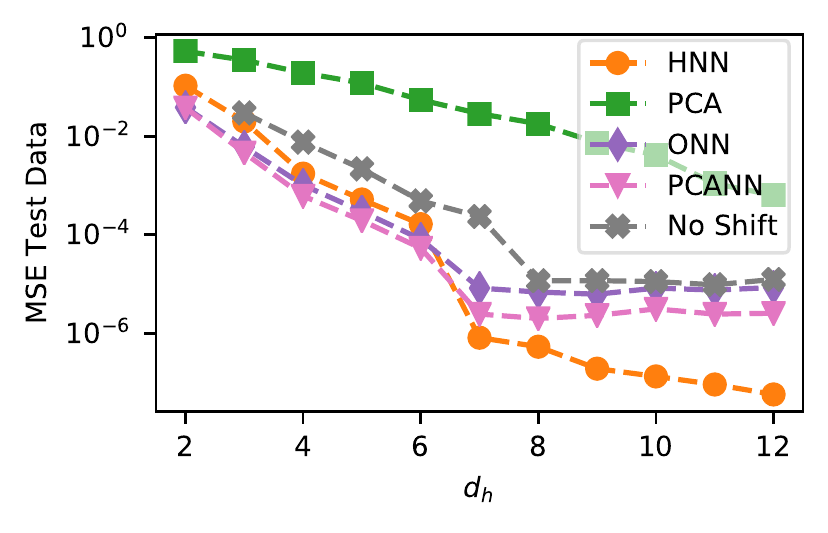}
	\captionsetup{justification=raggedright}
	\caption{MSE of test data for various $d_h$ for $L=22$. The legend is described in the text.}
	\label{fig:Autoencoder}
	\vspace{-5mm}
\end{figure} 

Autoencoders of the above structure, which we denote hybrid neural networks (HNN) were trained (i.e.~the functions $E$ and $D$ were determined) using the phase aligned data. At a given value of $d_h$, twenty HNNs (each initialized with different initial guesses for the weights), with $\alpha=1$ were trained for $1000$ epochs with an Adam optimizer using Keras \cite{chollet2015keras}. This process was repeated for a range of $d_h$. Results are reported for the model with the lowest MSE at each value of $d_h$. 

For comparison we trained three variations on the HNN to evaluate the effect of the linear projection ($P_{d_h}$), the PCA change of basis ($U\trans$), and phase alignment steps. In the first variation, denoted PCANN, we built a NN without the ``Trunc" and ``Expand" blocks in Fig. \ref{fig:Framework}, which corresponds to using the PCA basis, but using   $E$ and $D$ to learn the whole nonlinear coordinate tranformation rather than just the difference from PCA. The next variation builds upon the previous and removes the ``PCA" block in Figure \ref{fig:Framework}, which leaves it in the original basis, so we denote it ONN. Then, the last variation is to remove the phase shift (No Shift). Both the PCANN and the ONN are trained with the loss $L=||\hat{u}-\tilde{\hat{u}}||^2$, while the unsifted variation is trained with $L=||u-\tilde{u}||^2$. Hyperparameter tuning of the NN architectures was performed manually by varying width, depth, and activation functions. \ALrevise{All of these variations used the same architecture, shown in Table \ref{Table}, for functions $E$ and $D$.} 
%For comparison, we also trained two autoencoders \ALrevise{without the linear dimension reduction ($P_{d_h}$) using the} loss $L=||\hat{u}-\tilde{\hat{u}}||^2$. One of these autoencoders does not include the PCA coordinate transformations \ALrevise{(i.e. $E(\hat{u}(t))=h(t)$), which we denote as standard NN (SNN),} and the other with the PCA coordinate transformation \ALrevise{(i.e. $E(U\trans\hat{u}(t))=h(t)$), which we denote PCANN}. All NN used the same architecture for $D$ and $E$. The architectures for these NNs appear in Table \ref{Table}. \ALrevise{Hyperparameter tuning was performed manually by varying width, depth, and activation functions.} 

Figure \ref{fig:Autoencoder} shows the mean squared error (MSE) on a separate test dataset for the NN methods described above and PCA for $L=22$. At low $d_h$, the HNN, PCANN, and ONN all perform similarly, and in all three cases the MSE drops significantly at $d_h=7$. For the case of no shifting, the drop appears at $d_h=8$ because the continuous translation symmetry has not been factored out. All NNs perform orders of magnitude better than PCA. On continuing to increase $d_h$, the MSE for the HNN continues to improve while the others stagnate, because the HNN only needs correct coefficients of the less relevant higher PCA modes, while the other methods modify all of them. Notably, the abrupt drop in MSE at $d_h=7$ coincides with the true dimension $d_\IM$ of the attractor as found in \citep{Ding2016}. The remaining error for $d_h\geq 7$ for the HNN is small, at $O(10^{-7})$, which follows from the fact that at this dimension an exact coordinate transformation exists, so the remaining error is approximation error.

Having in hand the coordinate representation for points on $\IM$, we now use NNs to learn the dynamical system (``exact" reduced-order model) on the manifold, corresponding to the ``$\tau\neq0$" branch in Figure \ref{fig:Framework}. This approach will be denoted ``HNN ROM''.
%Training the autoencoder provided a means to map data to the inertial manifold coordinates with $E$ and back to the invariant space with $D$. The next challenge \sout{was}\MDGrevise{is} \sout{ estimating functions for evolving trajectories forward keeping the autoencoder functions static} \MDGrevise{predicting the dynamics on the inertial manifold}. 
We construct discrete time mappings 
\begin{equation}
	h(t+\hat{\tau})=F_h(h(t)),~~ \Delta\phi\equiv\phi(t+\hat{\tau})-\phi(t)=F_\phi(h(t)),\label{eq:dynamics}
\end{equation}
where $F_h$ and $F_\phi$ have the architectures shown in Table \ref{Table}. We use the symbol $\hat{\tau}$ instead of $\tau$ to emphasize that the discrete time mappings use in-slice time. We chose $\hat{\tau}=2$, which reproduces trajectories well, by allowing for the signal to change an appreciable amount, but not too much, in one time interval. Setting $\hat{\tau}$ much smaller or larger results in poor model predictions.

\begin{figure} 
	\includegraphics[trim=0 0 0 0,width=8.6 cm,clip]{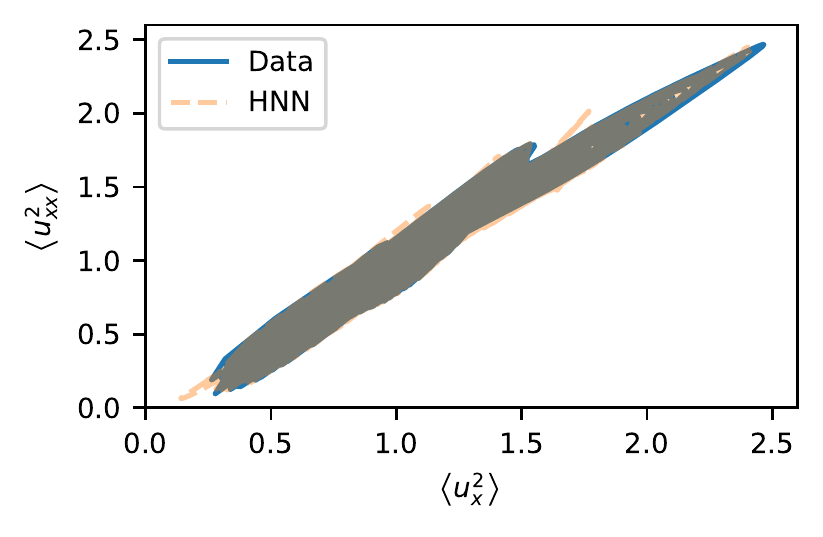}
	\captionsetup{justification=raggedright}
	\caption{$\mathcal{P}$ vs. $\varepsilon$ state-space projection for the data at $L=22$ and HNN ROM prediction with $d_h=7$.}
	\label{fig:PvD}
	\vspace{-5mm}
\end{figure}

Recall that the energy balance for the KSE requires that the production and dissipation rates $\mathcal{P}$ and $\varepsilon$ must balance on average. We incorporate this fact in the training of the dynamic models as follows. We compute the projection of the data onto $\mathcal{P}$ and $\varepsilon$, as shown in Figure \ref{fig:PvD}. The relation between $\mathcal{P}$ and $\varepsilon$ is narrowly distributed around the line $\mathcal{P}=\varepsilon$, with a sharp boundary, and we can find maximum and minimum dissipation rates $\varepsilon_{\text{max}}$ and $\varepsilon_{\text{min}}$ associated with a given value of $\mathcal{P}$. We then add a penalty for crossing this boundary to the loss function $L_F$ for $F_h$ and $F_\phi$, as follows: 
%
%\MDG{we don't have room for the details of constructing initial guesses etc.}
%%
%%\ALrevise{These mappings were trained in two steps. First, we separately trained $F_h$ on $\hat{u}_t$ and $F_\phi$ on $\Delta \phi$ to find a reasonable initial condition, then we trained both $F$ and $G$ together on $u_t$. This alleviated the difficulty of indirectly learning the phase shift by directly penalizing it when training the invariant model.} Additionally, to keep trajectories on the attractor, we added a penalty in the loss for energy balances that left the expected range of production vs dissipation values. 
%
%
%This was accomplished by calculating $P$ and $D$ at each time for a given solution of the KSE and then interpolating the boundary to find a maximum and minimum dissipation ($D_{\text{max}}$ and $D_{\text{min}}$) associated with a given value of $P$. Recall, $P$ and $D$ at a given point in time do not necessarily balance, so this approach was taken to penalize incorrect single time steps that deviate far from the expected range of values. Tildes over $P$ and $D$ indicate these values were calculated from $\tilde{u}$. The loss used for training in the invariant space is 
%\begin{multline} \label{eq:Loss2}
%L_2=||\hat{u}_{t+1}-\tilde{\hat{u}}_{t+1}||^2+||\Delta \phi-\Delta \tilde{\phi}||^2 \\
%+\beta \max(\max (0,\tilde{D}\!-\!D_{\text{max}}(\tilde{P})),D_{\text{min}}(\tilde{P})\!-\!\tilde{D}),
%\end{multline}
%and the final full-space loss is
\begin{multline} \label{eq:Loss3}
	L_F=||u(t+\hat{\tau})-\tilde{u}(t+\hat{\tau})||^2 \\
	+\beta \max(\max (0,\tilde{\varepsilon}\!-\!\varepsilon_{\text{max}}(\tilde{\mathcal{P}})),\varepsilon_{\text{min}}(\tilde{\mathcal{P}})\!-\!\tilde{\varepsilon}),
\end{multline}%\MDG{Add a few words about training? Value of $\beta$}
where $\tilde{\mathcal{P}}$ and $\tilde{\varepsilon}$ are calculated from $\tilde{u}$. We selected $\beta=0.1$ so the second term contributed the same order of error to the loss as the first term. For each $d_h$, the best dimension reduction model was chosen, and fifty time-evolution models were trained for 200 epochs. Results are reported for the best models, as determined at a given $d_h$ based on producing low errors in both short and long-time statistics.

\begin{figure} 
	\includegraphics[trim=0 0 0 0,width=8.6 cm,clip]{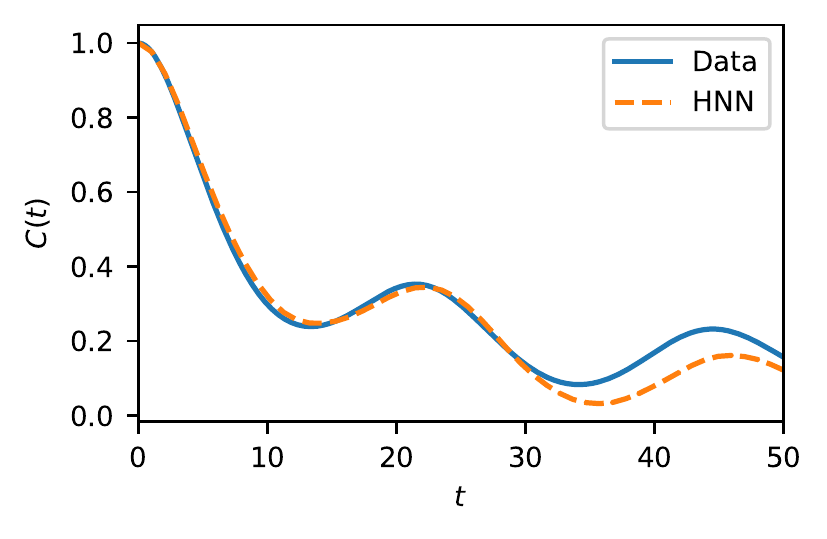}
	\captionsetup{justification=raggedright}
	\caption{Time-correlation function for the data at $L=22$, and HNN ROM prediction with $d_h=7$.}
	\label{fig:Temp}
	\vspace{-5mm}
\end{figure}
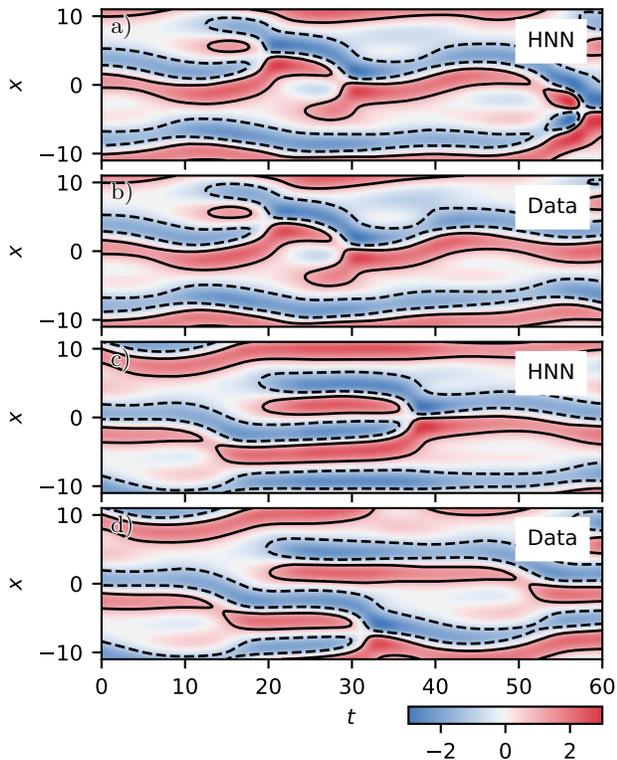
\begin{figure}
	\centering
	\captionsetup[subfigure]{labelformat=empty}
	\begin{subfigure}[b]{8.6 cm}
		\includegraphics[trim=0 0 0 0,width=\textwidth,clip]{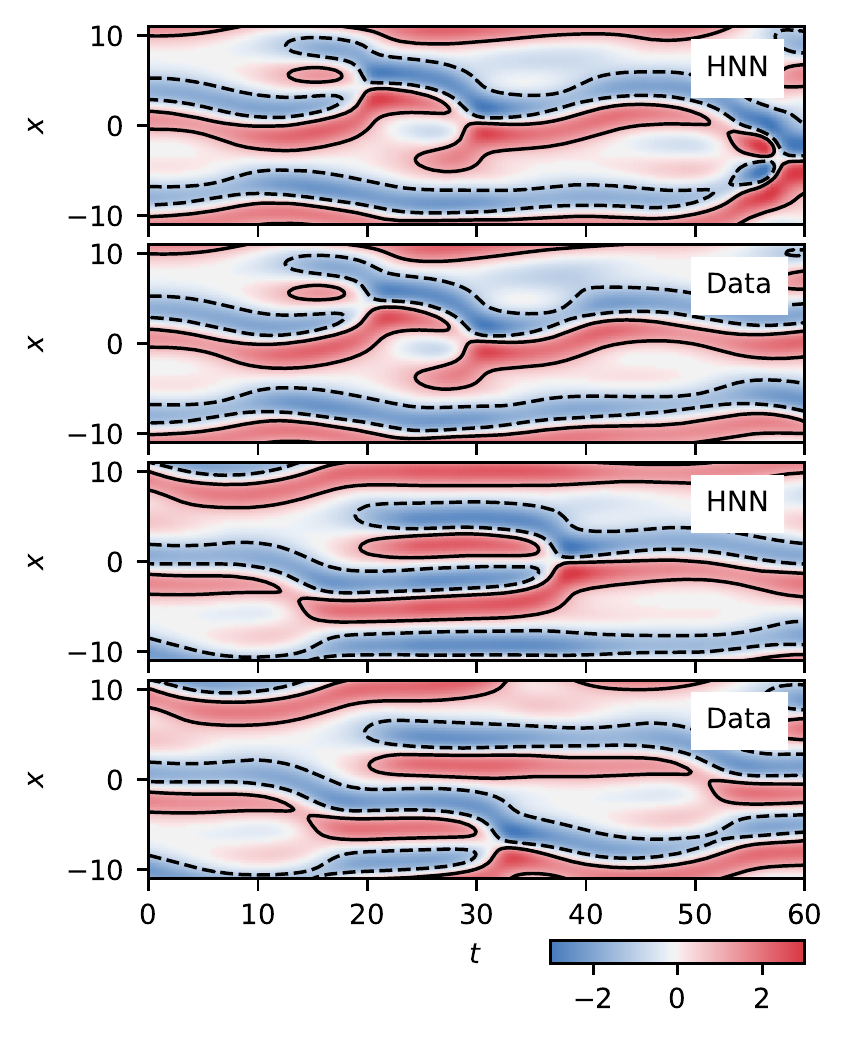}
		\begin{picture}(0,0)
		\put(-79,294){\contour{white}{ \textcolor{black}{a)}}}
		\put(-79,231){\contour{white}{ \textcolor{black}{b)}}}
		\put(-79,168){\contour{white}{ \textcolor{black}{c)}}}
		\put(-79,105){\contour{white}{ \textcolor{black}{d)}}}
		\end{picture}
		\caption{}
		\vspace{-10mm}
		\label{fig:Traja}
	\end{subfigure}
	\begin{subfigure}[b]{0\textwidth}\caption{}\vspace{-10mm}\label{fig:Trajb}\end{subfigure}\begin{subfigure}[b]{0\textwidth}\caption{}\vspace{-10mm}\label{fig:Trajc}\end{subfigure}\begin{subfigure}[b]{0\textwidth}\caption{}\vspace{-10mm}\label{fig:Trajd}\end{subfigure}    
	\vspace{-1.5\baselineskip}
	\captionsetup{justification=raggedright}
	\caption{(a) and (c) trajectories (color contours of $u(x,t)$ with solid lines at $u=1$ and dashed lines at $u=-1$) evolving from different initial conditions according to the HNN ROM; $L=22$. (c) and (d) true trajectories corresponding to (b) and (d).} %\ref{fig:Traja} and \ref{fig:Trajb} are two randomly selected example trajectories with correspond HNN predictions \ref{fig:Trajc} and \ref{fig:Trajd}.\MDG{rewrite this caption and the rest in the style of the caption to Figure 2. Figure captions are not narratives. You don't need to use full sentences}\MDG{flip so the NN predictions are on top}\MDG{time or in-slice time}}
	\label{fig:Traj} 
	\vspace{-5mm}
\end{figure}

%\MDG{we need to do some rewriting here to emphasize that (1) short-time predictions track ``exact" results for more than a Lyapunov time, which is very good performance for a model prediction of a chaotic system, and (2) the long-time dynamics stay on the attractor, as indicated by the PDFs that we show.}
To illustrate the performance of this approach, which we denote HNN ROM, on predicting dynamics, we first present short-time tracking results and then long-time statistics. All trajectories are evolved from a given initial condition $u(0)$ on the manifold, from which we find $\hat{u}(0)$ and $\phi(0)$ by phase alignment and then set $h(0)=\chi(\hat{u}(0))$. This initial condition in the manifold coordinates is evolved forward in in-slice time with Eq.~\ref{eq:dynamics}. For validating the performance of the short-time tracking we need a timescale for comparison. Here we consider the integral timescale $T_I=\int_0^\infty C(t)\;dt \approx 19$, where
\begin{equation*}
	C(t)=\frac{\langle u(0)u(t)\rangle}{\langle u(0)^2\rangle}
\end{equation*} 
is the temporal autocorrelation, and the Lyapunov time $T_L\approx 21$ \cite{Ding2016}. Figure \ref{fig:Temp} shows the temporal autocorrelation of the data and the HNN ROM at $d_h=d_\IM$ are in good agreement for $t\lesssim 30$. Likewise, typical trajectories show close tracking for 30 or more real time units. This comparison appears in Fig. \ref{fig:Traj}, where Fig. \ref{fig:Traja} and \ref{fig:Trajc} show the evolution of two initial conditions of test data using the dynamical system found with $d_h=d_\IM=7$, and the ``exact" results are shown in Fig. \ref{fig:Trajc} and \ref{fig:Trajd} obtained from solving the KSE. 
% The average distance between the HNN ROM and the data over an ensemble of $1\cdot10^5$ initial conditions is shown in \ref{fig:Dif}. It takes, on average, around 80 time units for the trajectories to be the same distance apart as randomly sampled data points. 
These results indicate predictive capability for time scales longer than $T_I$ or $T_L$, and thus represent very good performance for prediction of chaotic dynamics.

Next, we evaluate the ability of our dynamic model to reproduce key long-time statistical properties of the attractor, focusing on the quantities $u_x$ and $u_{xx}$ that determine the energy production and dissipation in the KSE.  We examine predictions both for $d_h=d_\IM$ and for values of $d_h$ either larger or smaller than $d_\IM$. The trajectories considered here cover approximately $9\cdot 10^3$ real time units.

\begin{figure} 
	\includegraphics[trim=0 0 0 0,width=8.6 cm,clip]{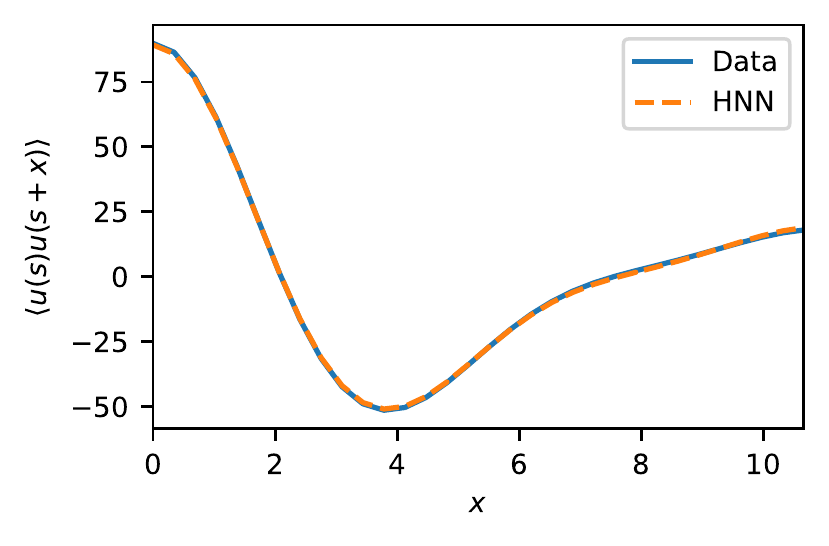}
	\captionsetup{justification=raggedright}
	\caption{Spatial correlation function for the data and HNN ROM prediction with $d_h=7$, $L=22$.}
	\label{fig:Spat}
	\vspace{-5mm}
\end{figure}

Figures \ref{fig:Spat} and \ref{fig:PvD}, respectively, show the spatial autocorrelation function (averaged over space and time) and the energy balance ($\mathcal{P}$ vs.~$\varepsilon$) of the HNN ROM for $d_h=d_\IM$ and for data, illustrating close agreement of these quantities. These statistics show that long-time trajectories do not diverge from the attractor and that the HNN ROM prediction stays within the envelope of the energy balance, which was the intent of the penalty in the loss for the time-evolution training, Eq.~\ref{eq:Loss3}. These predictions deteriorate when $d_h<d_\IM$.

%The spatial autocorrelation function (averaged over space and time) appears in Figure \ref{fig:Spat} showing very close agreement indicating similarly correlated signals. Figure \ref{fig:PvD} shows that the long-time HNN ROM prediction stays within the envelope of the energy balance, which was the intent of the penalty in the loss. } %we describe how the \ALrevise{attractor recreation of long-time} model predictions vary with $d_h$ and compare to predictions from linear dimension reduction with PCA and NNs for the dynamics, denoted PCA ROM. 

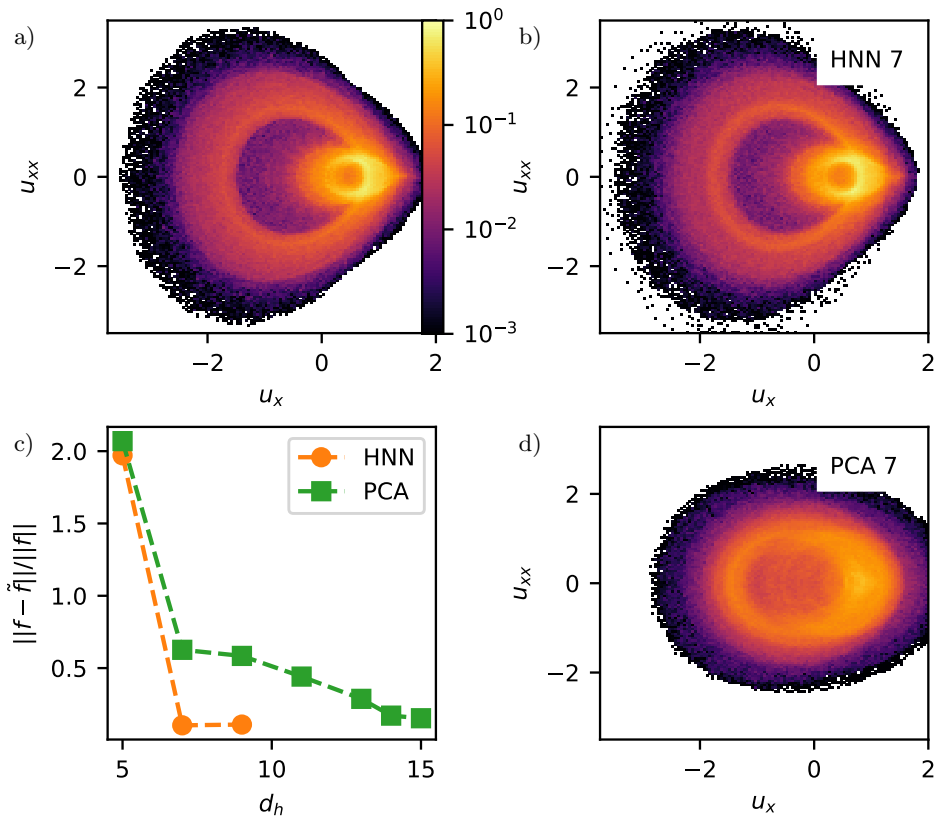
\begin{figure*}
	\centering
	\captionsetup[subfigure]{labelformat=empty}
	\begin{subfigure}[b]{12.9 cm}
		\includegraphics[trim=0 0 0 0,width=\textwidth,clip]{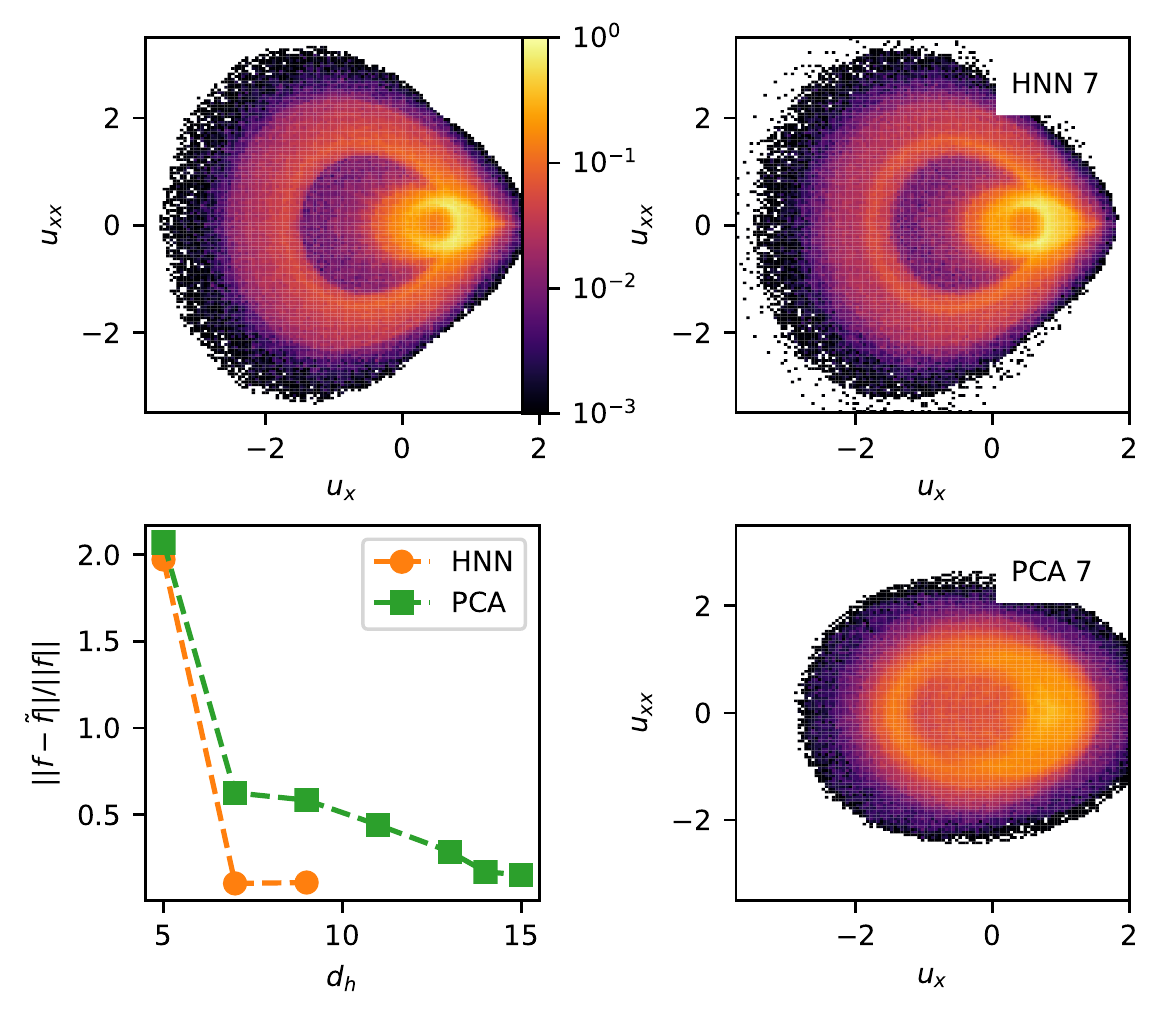}
		\begin{picture}(0,0)
		\put(-173,312){a)}
		\put(18,312){b)}
		\put(-173,158){c)}
		\put(18,158){d)}
		\end{picture}
		\caption{}
		\vspace{-10mm}
		\label{fig:PDF_Dat}
	\end{subfigure}
	\begin{subfigure}[b]{0\textwidth}\caption{}\vspace{-10mm}\label{fig:PDF_HL7}\end{subfigure}\begin{subfigure}[b]{0\textwidth}\caption{}\vspace{-10mm}\label{fig:PDF_Comp}\end{subfigure}\begin{subfigure}[b]{0\textwidth}\caption{}\vspace{-10mm}\label{fig:PDF_PCA7}\end{subfigure}	
	%\vspace{-1.5\baselineskip}
	\captionsetup{justification=raggedright}
	\caption{ (a) joint PDF of data, $L=22$, and (b) joint PDF of HNN ROM prediction, both plotted on a logarithmic scale. (c) Relative error in PDF vs dimension. \ALrevise{The PDF from data is denoted $f$, and that from the model prediction is $\tilde{f}$.}}%\MDG{make each plot 1 column wide and similarly for the rest of the paper}}% \ref{fig:1} is the joint PDF of $u_{xx}$ and $u_x$ for the KSE. HNN reconstructions appear in \ref{fig:2}-\ref{fig:4} for $d_h=5,7,9$ and PCA reconstructions appear in \ref{fig:6}-\ref{fig:8}. The difference between the reconstruction and \ref{fig:1} are summarized in \ref{fig:5}.\MDG{for f, are you showing a transient? If the attractor is a limit cycle, the pdf should just be a single closed curve.}}\label{fig:lab} 
	\vspace{-5mm}
\end{figure*}

A more detailed representation of the attractor is the joint probability density function (PDF) of the pointwise values of $u_x$ and $u_{xx}$.
%For these comparisons, we present the joint probability density function (PDF) of the pointwise values of $u_x$ and $u_{xx}$, because of their relevance for the overall energy balance. 
Figure \ref{fig:PDF_Dat} shows this PDF, on a log scale, as determined from the data. At  $d_h=d_\IM=7$, the HNN ROM prediction, Fig. \ref{fig:PDF_HL7}, is very close to the exact PDF. To highlight the effect of the nonlinear autoencoder, we also consider predictions with linear dimension reduction from PCA (i.e.~$E=D=0$), and NNs for the dynamics; we denote this approach PCA ROM. At the same dimension, Fig. \ref{fig:PDF_PCA7} shows that the PCA ROM prediction yields much poorer results. Figure \ref{fig:PDF_Comp} shows how the relative $L_2$ difference between the true PDF and the model predictions varies with $d_h$. For $d_h=5$, the predictions are poor for both cases, but for $d_h\geq 7$,the error is small and nearly unchanging for the HNN ROM case, which is unsurprising since $d_\IM=7$. On the other hand, it takes $d_h=14$ for the PCA ROM to yield a comparable model to the HNN ROM at $d_h=7$. This result might be expected based on Whitney's embedding theorem, which states that any manifold of dimension $k$ can be embedded in $\reals^{2k}$ \cite{Guillemin:2000ti}. 

%\begin{figure}
%	\centering
%	\captionsetup[subfigure]{labelformat=empty}
%	\begin{subfigure}[b]{8.6 cm}
%		\includegraphics[trim=0 0 0 0,width=\textwidth,clip]{pdf_figures/Linear}
%		\begin{picture}(0,0)
%		\put(-120,105){a)}
%		\put(5,105){b)}
%		\end{picture}
%		\caption{}
%		\vspace{-10mm}
%		\label{fig:Drop}
%	\end{subfigure}
%	\begin{subfigure}[b]{0\textwidth}\caption{}\vspace{-10mm}\label{fig:DvL}\end{subfigure}	\vspace{-1.5\baselineskip}
%	\captionsetup{justification=raggedright}
%	\caption{
%		(a)MSE vs the difference between $d_h$ and the predicted $d_\IM$. (b) Dimension $d_\IM+1$ of the unreduced IM for different domain sizes: \MDGrevise{``HNN" denotes the current results, ``PM" the number of physical modes from \cite{Yang2009}, and ``PM Fit" a linear fit of the PM curve.}} 
%	\vspace{-5mm}
%\end{figure}    

\begin{figure*}
	\centering
	\captionsetup[subfigure]{labelformat=empty}
	\begin{subfigure}[b]{17.2 cm}
		\includegraphics[trim=0 0 0 0,width=\textwidth,clip]{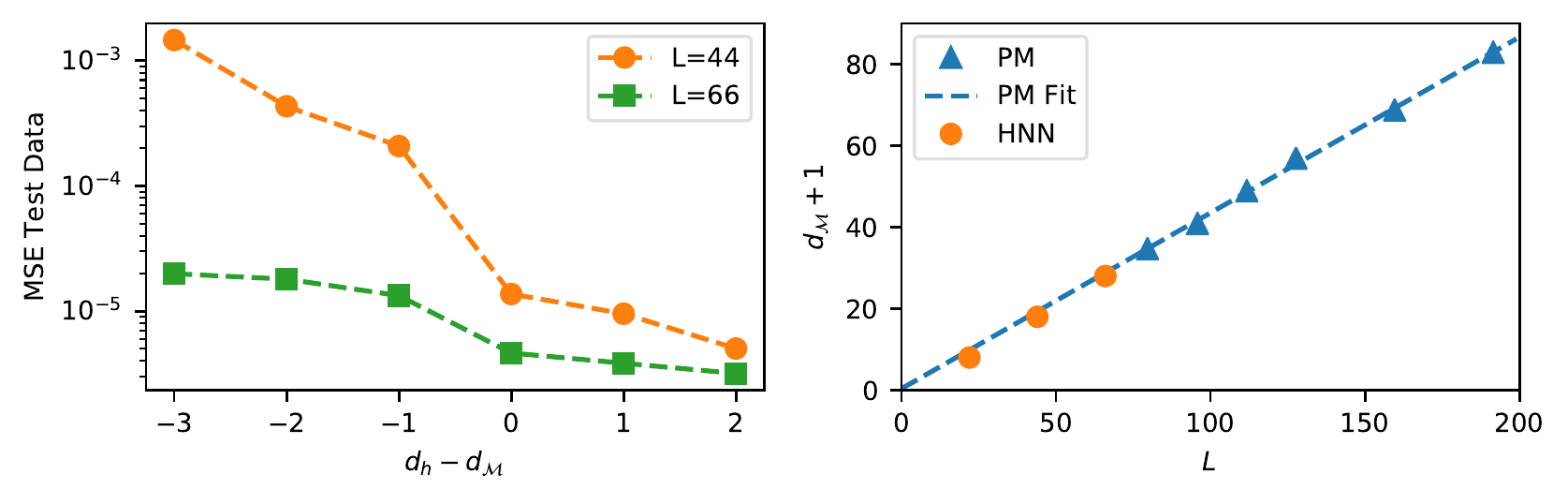}
		\begin{picture}(0,0)
		\put(-240,150){a)}
		\put(0,150){b)}
		\end{picture}
		\caption{}
		\vspace{-10mm}
		\label{fig:Drop}
	\end{subfigure}
	\begin{subfigure}[b]{0\textwidth}\caption{}\vspace{-10mm}\label{fig:DvL}\end{subfigure}
	%\vspace{-1.5\baselineskip}
	\captionsetup{justification=raggedright}
	\caption{ (a) MSE vs the difference between $d_h$ and the predicted $d_\IM$. (b) Dimension $d_\IM+1$ of the unreduced IM for different domain sizes: ``HNN" denotes the current results, ``PM" the number of physical modes from \cite{Yang2009}, and ``PM Fit" a linear fit of the PM curve.}%\MDG{make each plot 1 column wide and similarly for the rest of the paper}}% \ref{fig:1} is the joint PDF of $u_{xx}$ and $u_x$ for the KSE. HNN reconstructions appear in \ref{fig:2}-\ref{fig:4} for $d_h=5,7,9$ and PCA reconstructions appear in \ref{fig:6}-\ref{fig:8}. The difference between the reconstruction and \ref{fig:1} are summarized in \ref{fig:5}.\MDG{for f, are you showing a transient? If the attractor is a limit cycle, the pdf should just be a single closed curve.}}\label{fig:lab} 
	\vspace{-5mm}
\end{figure*}
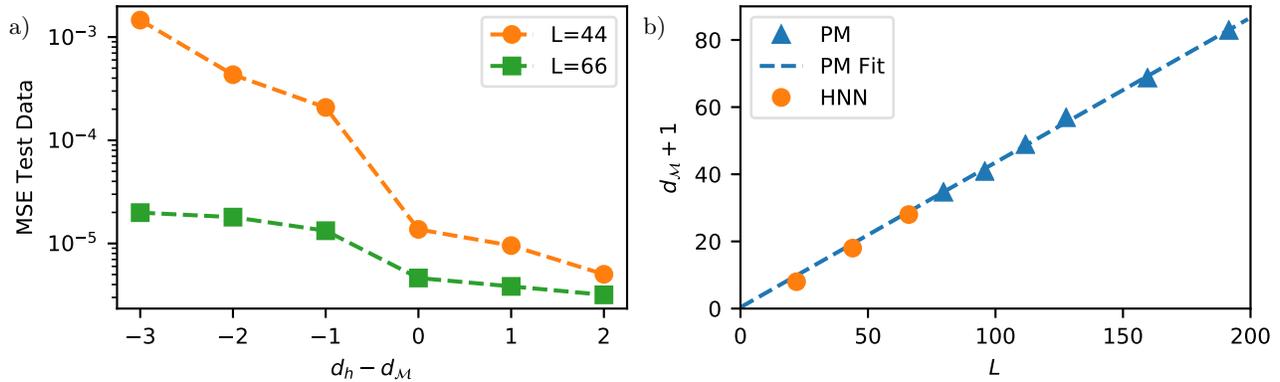

To investigate the generality of this method, we examine its performance for larger domains, $L=44$ and $L=66$. For $L=22$ there is one positive Lyapunov exponent \citep{Ding2016,Edson2019}, whereas the dynamics at $L=44$ and $L=66$ are more chaotic\MDGrevise{; at these values, Edson et al. \citep{Edson2019} report four and seven positive Lyapunov exponents, respectively.}  Data gathering and NN training were performed in the same way as for $L=22$. Initial trials for the $L=66$ showed poor results, so the capacity of the decoder was increased,as noted in Table \ref{Table}. Figure \ref{fig:Drop} shows the MSE on a test data set of HNNs with the lowest MSE at various $d_h$ for $L=44$ and $66$. The horizontal axis is centered around the $d_\IM$ of each domain size inferred from the drop in the MSE. This corresponds to $d_\IM=17$ and $d_\IM=27$ for $L=44$ and $L=66$. Increasing dimension still shows a distinct drop in MSE, however it becomes less substantial with increased domain size. The $L=66$ autoencoder performs better has lower MSE than $L=44$ because of the increased capacity.

%\begin{figure} 
%	\includegraphics[trim=0 0 0 0,width=8.6 cm,clip]{pdf_figures_final/LargeBoxMSEvdh}
%	\captionsetup{justification=raggedright}
%	\caption{MSE vs the difference between $d_h$ and the predicted $d_\IM$.}
%	\label{fig:Drop}
%	\vspace{-5mm}
%\end{figure}

The results at these domain sizes suggests that there is a linear scaling of $d_\IM$ with $L$. This observation agrees well with results of Yang et al. \citep{Yang2009} where they show the number of physical modes (PM) scales linearly with domain size. Figure \ref{fig:DvL} shows $d_\IM+1$ (the dimension in the unreduced state space) against the domain size for both our results and their results, along with the extrapolation of their results to smaller domains. The excellent agreement between these results provides additional computational evidence that the dimension of the IM for the KSE scales linearly with $L$.

%\begin{figure} 
%	\includegraphics[trim=0 0 0 0,width=8.6 cm,clip]{pdf_figures_final/DimvL}
%	\captionsetup{justification=raggedright}
%	\caption{Dimension $d_\IM+1$ of the unreduced IM for different domain sizes: ``HNN" denotes the current results, ``PM" the number of physical modes from \cite{Yang2009}, and ``PM Fit" a linear fit of the PM curve.}
%	\label{fig:DvL}
%	\vspace{-5mm}
%\end{figure}

\begin{figure*}
	\centering
	\captionsetup[subfigure]{labelformat=empty}
	\begin{subfigure}[b]{17.2 cm}
		\includegraphics[trim=0 0 0 0,width=\textwidth,clip]{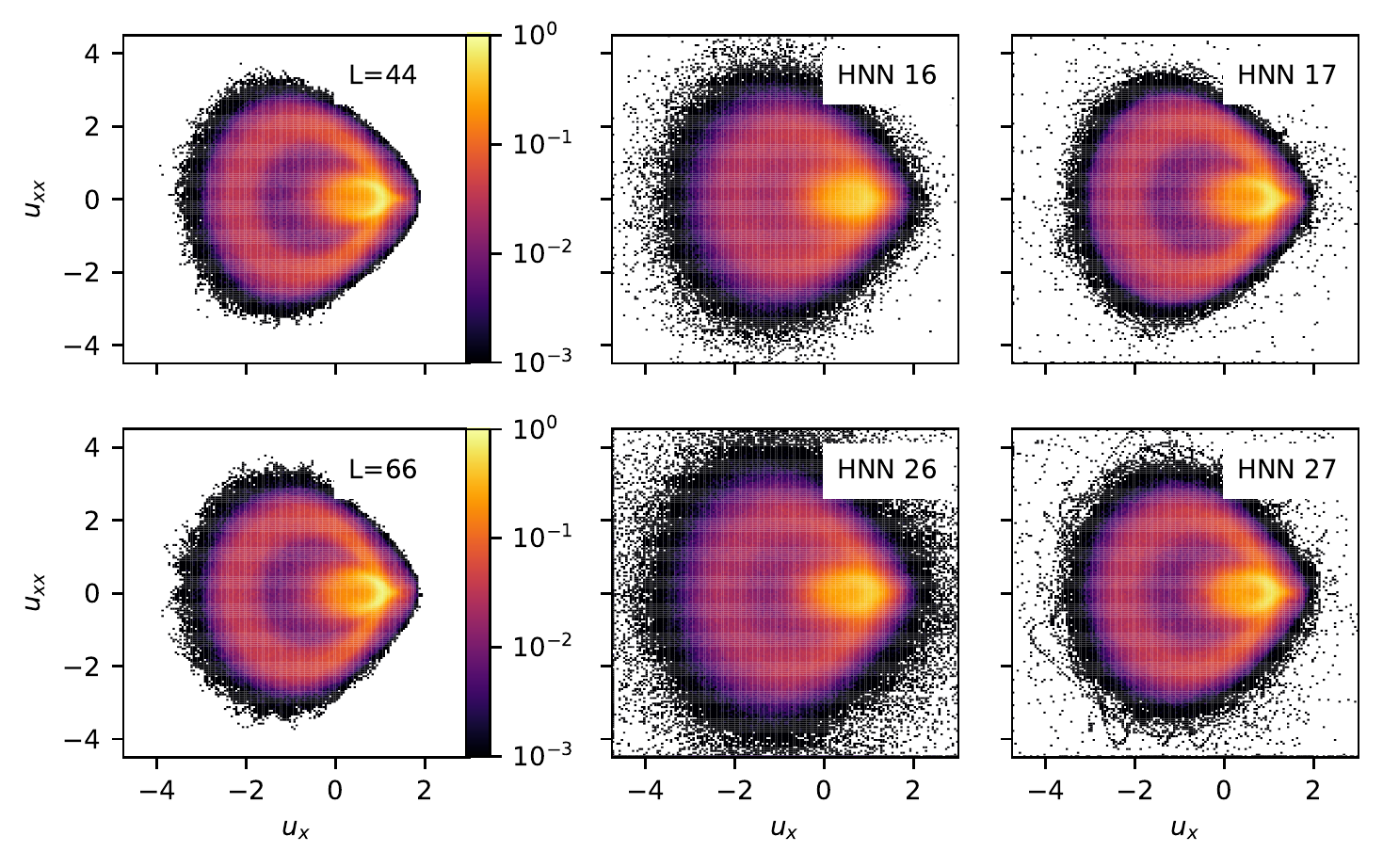}
		\begin{picture}(0,0)
		\put(-235,293){a)}
		\put(-45,293){b)}
		\put(98,293){c)}
		\put(-235,153){d)}
		\put(-45,153){e)}
		\put(98,153){f)}
		\end{picture}
		\caption{}
		\vspace{-10mm}
		\label{fig:PDF_Dat44}
	\end{subfigure}
	\begin{subfigure}[b]{0\textwidth}\caption{}\vspace{-10mm}\label{fig:PDF16}\end{subfigure}\begin{subfigure}[b]{0\textwidth}\caption{}\vspace{-10mm}\label{fig:PDF17}\end{subfigure}\begin{subfigure}[b]{0\textwidth}\caption{}\vspace{-10mm}\label{fig:PDF_Dat66}\end{subfigure}\begin{subfigure}[b]{0\textwidth}\caption{}\vspace{-10mm}\label{fig:PDF26}\end{subfigure}\begin{subfigure}[b]{0\textwidth}\caption{}\vspace{-10mm}\label{fig:PDF27}\end{subfigure}
	\captionsetup{justification=raggedright}
	\caption{ (a) and (d) joint PDF of data for $L=44,66$. (b), (c), (e), and (f) joint PDF of HNN ROM prediction at $d_\IM-1$ and $d_\IM$ for $L=44,66$.}
	\vspace{-5mm}
\end{figure*}
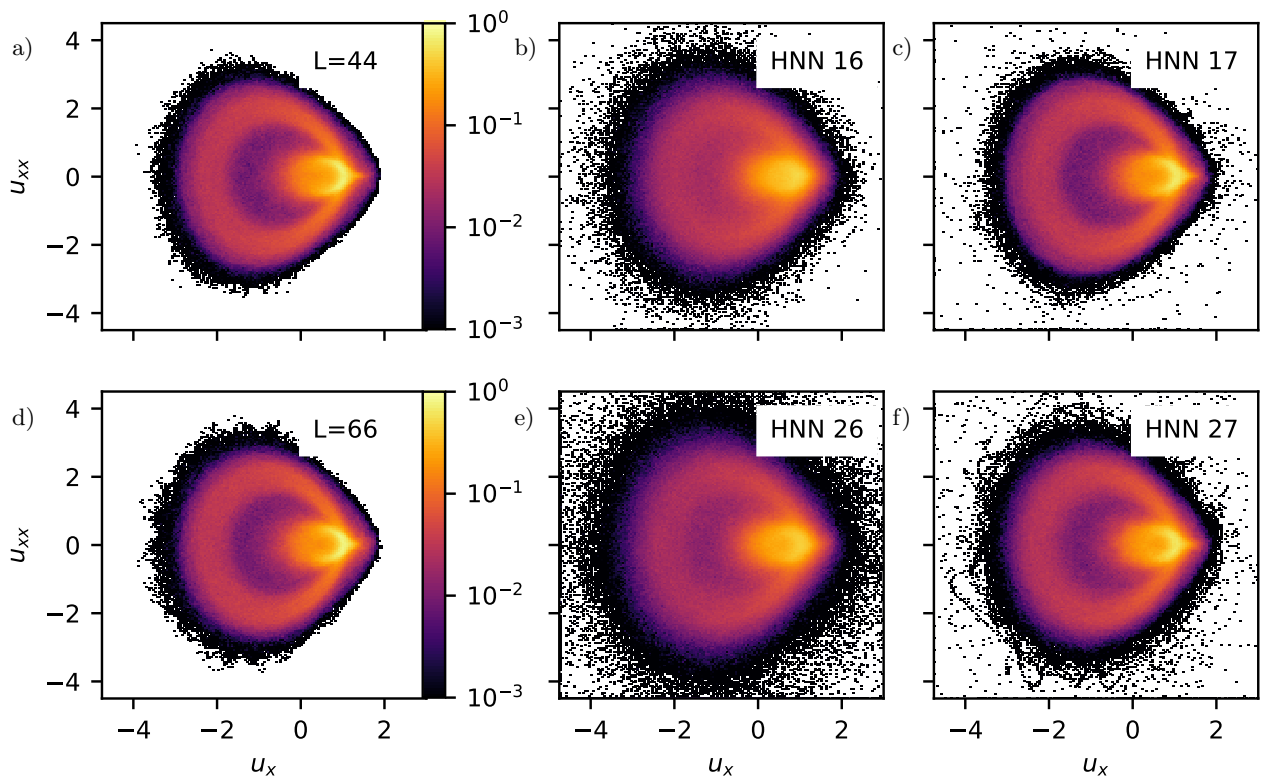   

Finally, we show the attractor recreation with the joint PDFs of $u_x$ and $u_{xx}$. Figure \ref{fig:PDF_Dat44} and \ref{fig:PDF_Dat66} show these for $L=44$ and $L=66$ for trajectories of approximately $8\cdot 10^3$ and $6\cdot 10^3$ real time units, respectively. Model predictions for $d_h=d_\IM-1$, Fig. \ref{fig:PDF16} and \ref{fig:PDF26}, and $d_Hd_\IM$ degrees of freedom, Fig. \ref{fig:PDF17} and \ref{fig:PDF27}, are shown for comparison. Unlike with $L=22$, where models with too few dimensions tended to land on periodic orbits, here, with more degrees of freedom, models with too few dimensions maintain chaos. However, the PDFs for $d_\IM-1$ are more diffuse, less accurately recreating the data. In both cases, when the models retain $d_\IM$ dimensions the joint PDF agrees well with the data, the main discrepancy being a broader tail of the model PDF, i.e.~ a higher, but still very low, probability of large excursions.

\section{Conclusion}
We have shown here a framework for data-driven ``exact'' reduction of a dynamical system onto a low dimensional invariant manifold and time-evolution on that manifold. Translation symmetry and energy conservation, two important features of many systems of interest, are incorporated naturally into the framework. By observing the model reduction error as a function of dimension, the dimension $d_\IM$ of the invariant manifold can be determined, and once $d_\IM$ is known, highly accurate model predictions can be obtained. In particular, key statistical quantities in a chaotic system can be well-approximated, indicating that the model dynamics capture the shape of the attractor. \ALrevise{In this work, the NNs can be trained on a single processor in a couple days, and time-evolution over $10^4$ time units takes only minutes. At present, it is difficult to predict how network size and training time will scale for more complex problems, especially given that it is not even known in general how $d_\IM$ scales.}  Extensions to systems with higher-dimensional dynamics are underway. Systematizing this method could provide a straightforward, data-driven means of approximating the dimension of manifolds and constructing reduced order models, a difficult task for high-dimensional chaotic systems like turbulence.

\begin{acknowledgments}
This work was supported by AFOSR  FA9550-18-1-0174 and ONR N00014-18-1-2865 (Vannevar Bush Faculty Fellowship). Code is available at https://github.com/alinot5/KSNN.git. 
\end{acknowledgments}

%\bibliography{library,papers-FDML,papers-ML}

\begin{thebibliography}{38}%
	\makeatletter
	\providecommand \@ifxundefined [1]{%
		\@ifx{#1\undefined}
	}%
	\providecommand \@ifnum [1]{%
		\ifnum #1\expandafter \@firstoftwo
		\else \expandafter \@secondoftwo
		\fi
	}%
	\providecommand \@ifx [1]{%
		\ifx #1\expandafter \@firstoftwo
		\else \expandafter \@secondoftwo
		\fi
	}%
	\providecommand \natexlab [1]{#1}%
	\providecommand \enquote  [1]{``#1''}%
	\providecommand \bibnamefont  [1]{#1}%
	\providecommand \bibfnamefont [1]{#1}%
	\providecommand \citenamefont [1]{#1}%
	\providecommand \href@noop [0]{\@secondoftwo}%
	\providecommand \href [0]{\begingroup \@sanitize@url \@href}%
	\providecommand \@href[1]{\@@startlink{#1}\@@href}%
	\providecommand \@@href[1]{\endgroup#1\@@endlink}%
	\providecommand \@sanitize@url [0]{\catcode `\\12\catcode `\$12\catcode
		`\&12\catcode `\#12\catcode `\^12\catcode `\_12\catcode `\%12\relax}%
	\providecommand \@@startlink[1]{}%
	\providecommand \@@endlink[0]{}%
	\providecommand \url  [0]{\begingroup\@sanitize@url \@url }%
	\providecommand \@url [1]{\endgroup\@href {#1}{\urlprefix }}%
	\providecommand \urlprefix  [0]{URL }%
	\providecommand \Eprint [0]{\href }%
	\providecommand \doibase [0]{https://doi.org/}%
	\providecommand \selectlanguage [0]{\@gobble}%
	\providecommand \bibinfo  [0]{\@secondoftwo}%
	\providecommand \bibfield  [0]{\@secondoftwo}%
	\providecommand \translation [1]{[#1]}%
	\providecommand \BibitemOpen [0]{}%
	\providecommand \bibitemStop [0]{}%
	\providecommand \bibitemNoStop [0]{.\EOS\space}%
	\providecommand \EOS [0]{\spacefactor3000\relax}%
	\providecommand \BibitemShut  [1]{\csname bibitem#1\endcsname}%
	\let\auto@bib@innerbib\@empty
	%</preamble>
	\bibitem [{\citenamefont {Hopf}(1948)}]{Hopf1948a}%
	\BibitemOpen
	\bibfield  {author} {\bibinfo {author} {\bibfnamefont {E.}~\bibnamefont
			{Hopf}},\ }\href {https://doi.org/10.1002/cpa.3160010401} {\bibfield
		{journal} {\bibinfo  {journal} {Communications on Pure and Applied
				Mathematics}\ }\textbf {\bibinfo {volume} {1}},\ \bibinfo {pages} {303}
		(\bibinfo {year} {1948})}\BibitemShut {NoStop}%
	\bibitem [{\citenamefont {Foias}\ \emph
		{et~al.}(1988{\natexlab{a}})\citenamefont {Foias}, \citenamefont
		{Nicolaenko}, \citenamefont {Sell},\ and\ \citenamefont
		{Temam}}]{Foias1988a}%
	\BibitemOpen
	\bibfield  {author} {\bibinfo {author} {\bibfnamefont {C.}~\bibnamefont
			{Foias}}, \bibinfo {author} {\bibfnamefont {B.}~\bibnamefont {Nicolaenko}},
		\bibinfo {author} {\bibfnamefont {G.~R.}\ \bibnamefont {Sell}},\ and\
		\bibinfo {author} {\bibfnamefont {R.}~\bibnamefont {Temam}},\ }\href@noop {}
	{\bibfield  {journal} {\bibinfo  {journal} {J. Math. Pure Appl.}\ }\textbf
		{\bibinfo {volume} {67}},\ \bibinfo {pages} {197} (\bibinfo {year}
		{1988}{\natexlab{a}})}\BibitemShut {NoStop}%
	\bibitem [{\citenamefont {Zelik}(2013)}]{Zelik2013}%
	\BibitemOpen
	\bibfield  {author} {\bibinfo {author} {\bibfnamefont {S.}~\bibnamefont
			{Zelik}},\ }\href {https://doi.org/10.1017/S0308210513000073} {\bibfield
		{journal} {\bibinfo  {journal} {Proceedings of the Royal Society of Edinburgh
				Section A: Mathematics}\ }\textbf {\bibinfo {volume} {144}},\ \bibinfo
		{pages} {1245} (\bibinfo {year} {2013})},\ \Eprint
	{https://arxiv.org/abs/1303.4457} {arXiv:1303.4457} \BibitemShut {NoStop}%
	\bibitem [{\citenamefont {Foias}\ \emph
		{et~al.}(1988{\natexlab{b}})\citenamefont {Foias}, \citenamefont {Sell},\
		and\ \citenamefont {Temam}}]{Foias1988}%
	\BibitemOpen
	\bibfield  {author} {\bibinfo {author} {\bibfnamefont {C.}~\bibnamefont
			{Foias}}, \bibinfo {author} {\bibfnamefont {G.~R.}\ \bibnamefont {Sell}},\
		and\ \bibinfo {author} {\bibfnamefont {R.}~\bibnamefont {Temam}},\ }\href
	{https://doi.org/10.1016/0022-0396(88)90110-6} {\bibfield  {journal}
		{\bibinfo  {journal} {Journal of Differential Equations}\ }\textbf {\bibinfo
			{volume} {73}},\ \bibinfo {pages} {309} (\bibinfo {year}
		{1988}{\natexlab{b}})}\BibitemShut {NoStop}%
	\bibitem [{\citenamefont {Graham}\ \emph {et~al.}(1993)\citenamefont {Graham},
		\citenamefont {Steen},\ and\ \citenamefont {Titi}}]{Graham:1993wv}%
	\BibitemOpen
	\bibfield  {author} {\bibinfo {author} {\bibfnamefont {M.~D.}\ \bibnamefont
			{Graham}}, \bibinfo {author} {\bibfnamefont {P.~H.}\ \bibnamefont {Steen}},\
		and\ \bibinfo {author} {\bibfnamefont {E.~S.}\ \bibnamefont {Titi}},\
	}\href@noop {} {\bibfield  {journal} {\bibinfo  {journal} {J Nonlinear Sci}\
		}\textbf {\bibinfo {volume} {3}},\ \bibinfo {pages} {153} (\bibinfo {year}
		{1993})}\BibitemShut {NoStop}%
	\bibitem [{\citenamefont {Hinton}\ and\ \citenamefont
		{Salakhutdinov}(2006)}]{Hinton:2006bg}%
	\BibitemOpen
	\bibfield  {author} {\bibinfo {author} {\bibfnamefont {G.~E.}\ \bibnamefont
			{Hinton}}\ and\ \bibinfo {author} {\bibfnamefont {R.~R.}\ \bibnamefont
			{Salakhutdinov}},\ }\href@noop {} {\bibfield  {journal} {\bibinfo  {journal}
			{Science}\ }\textbf {\bibinfo {volume} {313}},\ \bibinfo {pages} {504}
		(\bibinfo {year} {2006})}\BibitemShut {NoStop}%
	\bibitem [{\citenamefont {Goodfellow}\ \emph {et~al.}(2016)\citenamefont
		{Goodfellow}, \citenamefont {Bengio},\ and\ \citenamefont
		{Courville}}]{Goodfellow2016}%
	\BibitemOpen
	\bibfield  {author} {\bibinfo {author} {\bibfnamefont {I.}~\bibnamefont
			{Goodfellow}}, \bibinfo {author} {\bibfnamefont {Y.}~\bibnamefont {Bengio}},\
		and\ \bibinfo {author} {\bibfnamefont {A.}~\bibnamefont {Courville}},\
	}\href@noop {} {\emph {\bibinfo {title} {Deep Learning}}}\ (\bibinfo
	{publisher} {MIT Press},\ \bibinfo {year} {2016})\ \bibinfo {note}
	{\url{http://www.deeplearningbook.org}}\BibitemShut {NoStop}%
	\bibitem [{\citenamefont {Omata}\ and\ \citenamefont
		{Shirayama}(2019)}]{Omata:2019ho}%
	\BibitemOpen
	\bibfield  {author} {\bibinfo {author} {\bibfnamefont {N.}~\bibnamefont
			{Omata}}\ and\ \bibinfo {author} {\bibfnamefont {S.}~\bibnamefont
			{Shirayama}},\ }\href@noop {} {\bibfield  {journal} {\bibinfo  {journal} {AIP
				Advances}\ }\textbf {\bibinfo {volume} {9}},\ \bibinfo {pages} {015006}
		(\bibinfo {year} {2019})}\BibitemShut {NoStop}%
	\bibitem [{\citenamefont {Milano}\ and\ \citenamefont
		{Koumoutsakos}(2002)}]{Milano2002}%
	\BibitemOpen
	\bibfield  {author} {\bibinfo {author} {\bibfnamefont {M.}~\bibnamefont
			{Milano}}\ and\ \bibinfo {author} {\bibfnamefont {P.}~\bibnamefont
			{Koumoutsakos}},\ }\href {https://doi.org/10.1006/jcph.2002.7146} {\bibfield
		{journal} {\bibinfo  {journal} {Journal of Computational Physics}\ }\textbf
		{\bibinfo {volume} {182}},\ \bibinfo {pages} {1} (\bibinfo {year}
		{2002})}\BibitemShut {NoStop}%
	\bibitem [{\citenamefont {Otto}\ and\ \citenamefont {Rowley}(2019)}]{Otto2019}%
	\BibitemOpen
	\bibfield  {author} {\bibinfo {author} {\bibfnamefont {S.~E.}\ \bibnamefont
			{Otto}}\ and\ \bibinfo {author} {\bibfnamefont {C.~W.}\ \bibnamefont
			{Rowley}},\ }\href {https://doi.org/10.1137/18m1177846} {\bibfield  {journal}
		{\bibinfo  {journal} {{SIAM} Journal on Applied Dynamical Systems}\ }\textbf
		{\bibinfo {volume} {18}},\ \bibinfo {pages} {558} (\bibinfo {year}
		{2019})}\BibitemShut {NoStop}%
	\bibitem [{\citenamefont {Lusch}\ \emph {et~al.}(2018)\citenamefont {Lusch},
		\citenamefont {Kutz},\ and\ \citenamefont {Brunton}}]{Lusch:2018ex}%
	\BibitemOpen
	\bibfield  {author} {\bibinfo {author} {\bibfnamefont {B.}~\bibnamefont
			{Lusch}}, \bibinfo {author} {\bibfnamefont {J.~N.}\ \bibnamefont {Kutz}},\
		and\ \bibinfo {author} {\bibfnamefont {S.~L.}\ \bibnamefont {Brunton}},\
	}\href@noop {} {\bibfield  {journal} {\bibinfo  {journal} {Nat Comms}\
		}\textbf {\bibinfo {volume} {9}},\ \bibinfo {pages} {851} (\bibinfo {year}
		{2018})}\BibitemShut {NoStop}%
	\bibitem [{\citenamefont {Champion}\ \emph {et~al.}(2019)\citenamefont
		{Champion}, \citenamefont {Lusch}, \citenamefont {Kutz},\ and\ \citenamefont
		{Brunton}}]{Champion:2019vx}%
	\BibitemOpen
	\bibfield  {author} {\bibinfo {author} {\bibfnamefont {K.}~\bibnamefont
			{Champion}}, \bibinfo {author} {\bibfnamefont {B.}~\bibnamefont {Lusch}},
		\bibinfo {author} {\bibfnamefont {J.~N.}\ \bibnamefont {Kutz}},\ and\
		\bibinfo {author} {\bibfnamefont {S.~L.}\ \bibnamefont {Brunton}},\
	}\href@noop {} {\bibfield  {journal} {\bibinfo  {journal} {ArXiv}\ }
		(\bibinfo {year} {2019})},\ \Eprint {https://arxiv.org/abs/1904.02107v2}
	{1904.02107v2} \BibitemShut {NoStop}%
	\bibitem [{\citenamefont {Gonzalez}\ and\ \citenamefont
		{Balajewicz}(2018)}]{Gonzalez:2018tj}%
	\BibitemOpen
	\bibfield  {author} {\bibinfo {author} {\bibfnamefont {F.~J.}\ \bibnamefont
			{Gonzalez}}\ and\ \bibinfo {author} {\bibfnamefont {M.}~\bibnamefont
			{Balajewicz}},\ }\href@noop {} {\bibfield  {journal} {\bibinfo  {journal}
			{ArXiv}\ } (\bibinfo {year} {2018})},\ \Eprint
	{https://arxiv.org/abs/1808.01346v2} {1808.01346v2} \BibitemShut {NoStop}%
	\bibitem [{\citenamefont {Lee}\ and\ \citenamefont {Carlberg}(2019)}]{Lee2019}%
	\BibitemOpen
	\bibfield  {author} {\bibinfo {author} {\bibfnamefont {K.}~\bibnamefont
			{Lee}}\ and\ \bibinfo {author} {\bibfnamefont {K.~T.}\ \bibnamefont
			{Carlberg}},\ }\href {https://doi.org/10.1016/j.jcp.2019.108973} {\bibfield
		{journal} {\bibinfo  {journal} {Journal of Computational Physics}\ ,\
			\bibinfo {pages} {108973}} (\bibinfo {year} {2019})}\BibitemShut {NoStop}%
	\bibitem [{\citenamefont {Iten}\ \emph {et~al.}(2020)\citenamefont {Iten},
		\citenamefont {Metger}, \citenamefont {Wilming}, \citenamefont {del Rio},\
		and\ \citenamefont {Renner}}]{Iten2020}%
	\BibitemOpen
	\bibfield  {author} {\bibinfo {author} {\bibfnamefont {R.}~\bibnamefont
			{Iten}}, \bibinfo {author} {\bibfnamefont {T.}~\bibnamefont {Metger}},
		\bibinfo {author} {\bibfnamefont {H.}~\bibnamefont {Wilming}}, \bibinfo
		{author} {\bibfnamefont {L.}~\bibnamefont {del Rio}},\ and\ \bibinfo {author}
		{\bibfnamefont {R.}~\bibnamefont {Renner}},\ }\href
	{https://doi.org/10.1103/physrevlett.124.010508} {\bibfield  {journal}
		{\bibinfo  {journal} {Physical Review Letters}\ }\textbf {\bibinfo {volume}
			{124}},\ \bibinfo {pages} {10508} (\bibinfo {year} {2020})},\ \Eprint
	{https://arxiv.org/abs/1807.10300} {arXiv:1807.10300} \BibitemShut {NoStop}%
	\bibitem [{\citenamefont {Gonz{\'{a}}lez-Garc{\'{i}}a}\ \emph
		{et~al.}(1998)\citenamefont {Gonz{\'{a}}lez-Garc{\'{i}}a}, \citenamefont
		{Rico-Mart{\'{i}}nez},\ and\ \citenamefont
		{Kevrekidis}}]{Gonzalez-Garcia1998}%
	\BibitemOpen
	\bibfield  {author} {\bibinfo {author} {\bibfnamefont {R.}~\bibnamefont
			{Gonz{\'{a}}lez-Garc{\'{i}}a}}, \bibinfo {author} {\bibfnamefont
			{R.}~\bibnamefont {Rico-Mart{\'{i}}nez}},\ and\ \bibinfo {author}
		{\bibfnamefont {I.~G.}\ \bibnamefont {Kevrekidis}},\ }\bibfield  {journal}
	{\bibinfo  {journal} {Computers and Chemical Engineering}\ }\textbf {\bibinfo
		{volume} {22}},\ \href {https://doi.org/10.1016/s0098-1354(98)00191-4}
	{10.1016/s0098-1354(98)00191-4} (\bibinfo {year} {1998})\BibitemShut
	{NoStop}%
	\bibitem [{\citenamefont {Pathak}\ \emph {et~al.}(2018)\citenamefont {Pathak},
		\citenamefont {Hunt}, \citenamefont {Girvan}, \citenamefont {Lu},\ and\
		\citenamefont {Ott}}]{Pathak:2018dg}%
	\BibitemOpen
	\bibfield  {author} {\bibinfo {author} {\bibfnamefont {J.}~\bibnamefont
			{Pathak}}, \bibinfo {author} {\bibfnamefont {B.}~\bibnamefont {Hunt}},
		\bibinfo {author} {\bibfnamefont {M.}~\bibnamefont {Girvan}}, \bibinfo
		{author} {\bibfnamefont {Z.}~\bibnamefont {Lu}},\ and\ \bibinfo {author}
		{\bibfnamefont {E.}~\bibnamefont {Ott}},\ }\href@noop {} {\bibfield
		{journal} {\bibinfo  {journal} {Phys. Rev. Lett.}\ }\textbf {\bibinfo
			{volume} {120}},\ \bibinfo {pages} {024102} (\bibinfo {year}
		{2018})}\BibitemShut {NoStop}%
	\bibitem [{\citenamefont {Lui}\ and\ \citenamefont {Wolf}(2019)}]{Lui2019}%
	\BibitemOpen
	\bibfield  {author} {\bibinfo {author} {\bibfnamefont {H.~F.}\ \bibnamefont
			{Lui}}\ and\ \bibinfo {author} {\bibfnamefont {W.~R.}\ \bibnamefont {Wolf}},\
	}\href {https://doi.org/10.1017/jfm.2019.358} {\bibfield  {journal} {\bibinfo
			{journal} {Journal of Fluid Mechanics}\ }\textbf {\bibinfo {volume} {872}},\
		\bibinfo {pages} {963} (\bibinfo {year} {2019})}\BibitemShut {NoStop}%
	\bibitem [{\citenamefont {Vlachas}\ \emph {et~al.}(2018)\citenamefont
		{Vlachas}, \citenamefont {Byeon}, \citenamefont {Wan}, \citenamefont
		{Sapsis},\ and\ \citenamefont {Koumoutsakos}}]{Vlachas:2018gc}%
	\BibitemOpen
	\bibfield  {author} {\bibinfo {author} {\bibfnamefont {P.~R.}\ \bibnamefont
			{Vlachas}}, \bibinfo {author} {\bibfnamefont {W.}~\bibnamefont {Byeon}},
		\bibinfo {author} {\bibfnamefont {Z.~Y.}\ \bibnamefont {Wan}}, \bibinfo
		{author} {\bibfnamefont {T.~P.}\ \bibnamefont {Sapsis}},\ and\ \bibinfo
		{author} {\bibfnamefont {P.}~\bibnamefont {Koumoutsakos}},\ }\href@noop {}
	{\bibfield  {journal} {\bibinfo  {journal} {Proceedings of the Royal Society
				A: Mathematical, Physical and Engineering Sciences}\ }\textbf {\bibinfo
			{volume} {474}},\ \bibinfo {pages} {20170844} (\bibinfo {year}
		{2018})}\BibitemShut {NoStop}%
	\bibitem [{\citenamefont {Wan}\ \emph {et~al.}(2018)\citenamefont {Wan},
		\citenamefont {Vlachas}, \citenamefont {Koumoutsakos},\ and\ \citenamefont
		{Sapsis}}]{Wan:2018ga}%
	\BibitemOpen
	\bibfield  {author} {\bibinfo {author} {\bibfnamefont {Z.~Y.}\ \bibnamefont
			{Wan}}, \bibinfo {author} {\bibfnamefont {P.}~\bibnamefont {Vlachas}},
		\bibinfo {author} {\bibfnamefont {P.}~\bibnamefont {Koumoutsakos}},\ and\
		\bibinfo {author} {\bibfnamefont {T.}~\bibnamefont {Sapsis}},\ }\href@noop {}
	{\bibfield  {journal} {\bibinfo  {journal} {PLoS ONE}\ }\textbf {\bibinfo
			{volume} {13}},\ \bibinfo {pages} {e0197704} (\bibinfo {year}
		{2018})}\BibitemShut {NoStop}%
	\bibitem [{\citenamefont {Titi}(1990)}]{Titi1990}%
	\BibitemOpen
	\bibfield  {author} {\bibinfo {author} {\bibfnamefont {E.~S.}\ \bibnamefont
			{Titi}},\ }\href {https://doi.org/10.1016/0022-247X(90)90061-J} {\bibfield
		{journal} {\bibinfo  {journal} {Journal of Mathematical Analysis and
				Applications}\ }\textbf {\bibinfo {volume} {149}},\ \bibinfo {pages} {540}
		(\bibinfo {year} {1990})}\BibitemShut {NoStop}%
	\bibitem [{\citenamefont {Foias}\ \emph
		{et~al.}(1988{\natexlab{c}})\citenamefont {Foias}, \citenamefont {Jolly},
		\citenamefont {Kevrekidis}, \citenamefont {Sell},\ and\ \citenamefont
		{Titi}}]{Foias1988b}%
	\BibitemOpen
	\bibfield  {author} {\bibinfo {author} {\bibfnamefont {C.}~\bibnamefont
			{Foias}}, \bibinfo {author} {\bibfnamefont {M.~S.}\ \bibnamefont {Jolly}},
		\bibinfo {author} {\bibfnamefont {I.~G.}\ \bibnamefont {Kevrekidis}},
		\bibinfo {author} {\bibfnamefont {G.~R.}\ \bibnamefont {Sell}},\ and\
		\bibinfo {author} {\bibfnamefont {E.~S.}\ \bibnamefont {Titi}},\ }\href
	{https://doi.org/10.1016/0375-9601(88)90295-2} {\bibfield  {journal}
		{\bibinfo  {journal} {Physics Letters A}\ }\textbf {\bibinfo {volume}
			{131}},\ \bibinfo {pages} {433} (\bibinfo {year}
		{1988}{\natexlab{c}})}\BibitemShut {NoStop}%
	\bibitem [{\citenamefont {Jolly}\ \emph {et~al.}(1990)\citenamefont {Jolly},
		\citenamefont {Kevrekidis},\ and\ \citenamefont {Titi}}]{Jolly1990}%
	\BibitemOpen
	\bibfield  {author} {\bibinfo {author} {\bibfnamefont {M.~S.}\ \bibnamefont
			{Jolly}}, \bibinfo {author} {\bibfnamefont {I.~G.}\ \bibnamefont
			{Kevrekidis}},\ and\ \bibinfo {author} {\bibfnamefont {E.~S.}\ \bibnamefont
			{Titi}},\ }\href@noop {} {\bibfield  {journal} {\bibinfo  {journal} {Physica
				D: Nonlinear Phenomena}\ }\textbf {\bibinfo {volume} {44}} (\bibinfo {year}
		{1990})}\BibitemShut {NoStop}%
	\bibitem [{\citenamefont {Matthies}\ and\ \citenamefont
		{Meyer}(2003)}]{Matthies2003}%
	\BibitemOpen
	\bibfield  {author} {\bibinfo {author} {\bibfnamefont {H.~G.}\ \bibnamefont
			{Matthies}}\ and\ \bibinfo {author} {\bibfnamefont {M.}~\bibnamefont
			{Meyer}},\ }\href {https://doi.org/10.1016/S0045-7949(03)00042-7} {\bibfield
		{journal} {\bibinfo  {journal} {Computers and Structures}\ }\textbf {\bibinfo
			{volume} {81}},\ \bibinfo {pages} {1277} (\bibinfo {year}
		{2003})}\BibitemShut {NoStop}%
	\bibitem [{\citenamefont {Kuptsov}\ and\ \citenamefont
		{Kuptsova}(2019)}]{Kuptsov2019}%
	\BibitemOpen
	\bibfield  {author} {\bibinfo {author} {\bibfnamefont {P.~V.}\ \bibnamefont
			{Kuptsov}}\ and\ \bibinfo {author} {\bibfnamefont {A.~V.}\ \bibnamefont
			{Kuptsova}},\ }in\ \href {https://doi.org/10.1117/12.2523235} {\emph
		{\bibinfo {booktitle} {Saratov Fall Meeting 2018: Computations and Data
				Analysis: from Nanoscale Tools to Brain Functions}}},\ \bibinfo {editor}
	{edited by\ \bibinfo {editor} {\bibfnamefont {D.~E.}\ \bibnamefont
			{Postnov}}}\ (\bibinfo  {publisher} {{SPIE}},\ \bibinfo {year}
	{2019})\BibitemShut {NoStop}%
	\bibitem [{\citenamefont {Yang}\ \emph {et~al.}(2009)\citenamefont {Yang},
		\citenamefont {Takeuchi}, \citenamefont {Ginelli}, \citenamefont {Chat\'e},\
		and\ \citenamefont {Radons}}]{Yang2009}%
	\BibitemOpen
	\bibfield  {author} {\bibinfo {author} {\bibfnamefont {H.~L.}\ \bibnamefont
			{Yang}}, \bibinfo {author} {\bibfnamefont {K.~A.}\ \bibnamefont {Takeuchi}},
		\bibinfo {author} {\bibfnamefont {F.}~\bibnamefont {Ginelli}}, \bibinfo
		{author} {\bibfnamefont {H.}~\bibnamefont {Chat\'e}},\ and\ \bibinfo {author}
		{\bibfnamefont {G.}~\bibnamefont {Radons}},\ }\href
	{https://doi.org/10.1103/PhysRevLett.102.074102} {\bibfield  {journal}
		{\bibinfo  {journal} {Phys. Rev. Lett.}\ }\textbf {\bibinfo {volume} {102}},\
		\bibinfo {pages} {074102} (\bibinfo {year} {2009})}\BibitemShut {NoStop}%
	\bibitem [{\citenamefont {Ding}\ \emph {et~al.}(2016)\citenamefont {Ding},
		\citenamefont {Chat\'e}, \citenamefont {Cvitanovi\ifmmode~\acute{c}\else
			\'{c}\fi{}}, \citenamefont {Siminos},\ and\ \citenamefont
		{Takeuchi}}]{Ding2016}%
	\BibitemOpen
	\bibfield  {author} {\bibinfo {author} {\bibfnamefont {X.}~\bibnamefont
			{Ding}}, \bibinfo {author} {\bibfnamefont {H.}~\bibnamefont {Chat\'e}},
		\bibinfo {author} {\bibfnamefont {P.}~\bibnamefont
			{Cvitanovi\ifmmode~\acute{c}\else \'{c}\fi{}}}, \bibinfo {author}
		{\bibfnamefont {E.}~\bibnamefont {Siminos}},\ and\ \bibinfo {author}
		{\bibfnamefont {K.~A.}\ \bibnamefont {Takeuchi}},\ }\href
	{https://doi.org/10.1103/PhysRevLett.117.024101} {\bibfield  {journal}
		{\bibinfo  {journal} {Phys. Rev. Lett.}\ }\textbf {\bibinfo {volume} {117}},\
		\bibinfo {pages} {024101} (\bibinfo {year} {2016})}\BibitemShut {NoStop}%
	\bibitem [{\citenamefont {Kassam}\ and\ \citenamefont
		{Trefethen}(2005)}]{Kassam2005}%
	\BibitemOpen
	\bibfield  {author} {\bibinfo {author} {\bibfnamefont {A.-K.}\ \bibnamefont
			{Kassam}}\ and\ \bibinfo {author} {\bibfnamefont {L.~N.}\ \bibnamefont
			{Trefethen}},\ }\href {https://doi.org/10.1137/s1064827502410633} {\bibfield
		{journal} {\bibinfo  {journal} {{SIAM} Journal on Scientific Computing}\
		}\textbf {\bibinfo {volume} {26}},\ \bibinfo {pages} {1214} (\bibinfo {year}
		{2005})}\BibitemShut {NoStop}%
	\bibitem [{\citenamefont {Cvitanovi{\'c}}\ \emph {et~al.}(2016)\citenamefont
		{Cvitanovi{\'c}}, \citenamefont {Artuso}, \citenamefont {Mainieri},
		\citenamefont {Tanner},\ and\ \citenamefont {Vattay}}]{ChaosBook}%
	\BibitemOpen
	\bibfield  {author} {\bibinfo {author} {\bibfnamefont {P.}~\bibnamefont
			{Cvitanovi{\'c}}}, \bibinfo {author} {\bibfnamefont {R.}~\bibnamefont
			{Artuso}}, \bibinfo {author} {\bibfnamefont {R.}~\bibnamefont {Mainieri}},
		\bibinfo {author} {\bibfnamefont {G.}~\bibnamefont {Tanner}},\ and\ \bibinfo
		{author} {\bibfnamefont {G.}~\bibnamefont {Vattay}},\ }\href
	{http://ChaosBook.org/} {\emph {\bibinfo {title} {Chaos: Classical and
				Quantum}}}\ (\bibinfo  {publisher} {Niels Bohr Inst.},\ \bibinfo {address}
	{Copenhagen},\ \bibinfo {year} {2016})\BibitemShut {NoStop}%
	\bibitem [{\citenamefont {Budanur}\ \emph
		{et~al.}(2015{\natexlab{a}})\citenamefont {Budanur}, \citenamefont
		{Borrero-Echeverry},\ and\ \citenamefont {Cvitanovi{\'{c}}}}]{Budanur2015b}%
	\BibitemOpen
	\bibfield  {author} {\bibinfo {author} {\bibfnamefont {N.~B.}\ \bibnamefont
			{Budanur}}, \bibinfo {author} {\bibfnamefont {D.}~\bibnamefont
			{Borrero-Echeverry}},\ and\ \bibinfo {author} {\bibfnamefont
			{P.}~\bibnamefont {Cvitanovi{\'{c}}}},\ }\bibfield  {journal} {\bibinfo
		{journal} {Chaos}\ }\textbf {\bibinfo {volume} {25}},\ \href
	{https://doi.org/10.1063/1.4923742} {10.1063/1.4923742} (\bibinfo {year}
	{2015}{\natexlab{a}})\BibitemShut {NoStop}%
	\bibitem [{\citenamefont {Budanur}\ \emph
		{et~al.}(2015{\natexlab{b}})\citenamefont {Budanur}, \citenamefont
		{Cvitanovi\ifmmode~\acute{c}\else \'{c}\fi{}}, \citenamefont {Davidchack},\
		and\ \citenamefont {Siminos}}]{Budanur2015}%
	\BibitemOpen
	\bibfield  {author} {\bibinfo {author} {\bibfnamefont {N.~B.}\ \bibnamefont
			{Budanur}}, \bibinfo {author} {\bibfnamefont {P.}~\bibnamefont
			{Cvitanovi\ifmmode~\acute{c}\else \'{c}\fi{}}}, \bibinfo {author}
		{\bibfnamefont {R.~L.}\ \bibnamefont {Davidchack}},\ and\ \bibinfo {author}
		{\bibfnamefont {E.}~\bibnamefont {Siminos}},\ }\href
	{https://doi.org/10.1103/PhysRevLett.114.084102} {\bibfield  {journal}
		{\bibinfo  {journal} {Phys. Rev. Lett.}\ }\textbf {\bibinfo {volume} {114}},\
		\bibinfo {pages} {084102} (\bibinfo {year} {2015}{\natexlab{b}})}\BibitemShut
	{NoStop}%
	\bibitem [{\citenamefont {Willis}\ \emph {et~al.}(2013)\citenamefont {Willis},
		\citenamefont {Cvitanovi{\'{c}}},\ and\ \citenamefont {Avila}}]{Willis2013}%
	\BibitemOpen
	\bibfield  {author} {\bibinfo {author} {\bibfnamefont {A.~P.}\ \bibnamefont
			{Willis}}, \bibinfo {author} {\bibfnamefont {P.}~\bibnamefont
			{Cvitanovi{\'{c}}}},\ and\ \bibinfo {author} {\bibfnamefont {M.}~\bibnamefont
			{Avila}},\ }\href {https://doi.org/10.1017/jfm.2013.75} {\bibfield  {journal}
		{\bibinfo  {journal} {Journal of Fluid Mechanics}\ }\textbf {\bibinfo
			{volume} {721}},\ \bibinfo {pages} {514} (\bibinfo {year} {2013})},\ \Eprint
	{https://arxiv.org/abs/arXiv:1203.3701v1} {arXiv:arXiv:1203.3701v1}
	\BibitemShut {NoStop}%
	\bibitem [{\citenamefont {Strang}(2019)}]{Strang2019}%
	\BibitemOpen
	\bibfield  {author} {\bibinfo {author} {\bibfnamefont {G.}~\bibnamefont
			{Strang}},\ }\href@noop {} {\emph {\bibinfo {title} {Linear algebra and
				learning from data}}}\ (\bibinfo  {publisher} {Wellesley-Cambridge Press},\
	\bibinfo {year} {2019})\BibitemShut {NoStop}%
	\bibitem [{\citenamefont {van~der Maaten}\ and\ \citenamefont
		{Hinton}(2008)}]{vanDerMaaten2008}%
	\BibitemOpen
	\bibfield  {author} {\bibinfo {author} {\bibfnamefont {L.}~\bibnamefont
			{van~der Maaten}}\ and\ \bibinfo {author} {\bibfnamefont {G.}~\bibnamefont
			{Hinton}},\ }\href {http://www.jmlr.org/papers/v9/vandermaaten08a.html}
	{\bibfield  {journal} {\bibinfo  {journal} {Journal of Machine Learning
				Research}\ }\textbf {\bibinfo {volume} {9}},\ \bibinfo {pages} {2579}
		(\bibinfo {year} {2008})}\BibitemShut {NoStop}%
	\bibitem [{\citenamefont {Van Der~Maaten}\ \emph {et~al.}(2009)\citenamefont
		{Van Der~Maaten}, \citenamefont {Postma},\ and\ \citenamefont {Van~den
			Herik}}]{vanDerMaaten2009}%
	\BibitemOpen
	\bibfield  {author} {\bibinfo {author} {\bibfnamefont {L.}~\bibnamefont {Van
				Der~Maaten}}, \bibinfo {author} {\bibfnamefont {E.}~\bibnamefont {Postma}},\
		and\ \bibinfo {author} {\bibfnamefont {J.}~\bibnamefont {Van~den Herik}},\
	}\href@noop {} {\bibfield  {journal} {\bibinfo  {journal} {J Mach Learn Res}\
		}\textbf {\bibinfo {volume} {10}},\ \bibinfo {pages} {66} (\bibinfo {year}
		{2009})}\BibitemShut {NoStop}%
	\bibitem [{\citenamefont {Chollet}\ \emph {et~al.}(2015)\citenamefont {Chollet}
		\emph {et~al.}}]{chollet2015keras}%
	\BibitemOpen
	\bibfield  {author} {\bibinfo {author} {\bibfnamefont {F.}~\bibnamefont
			{Chollet}} \emph {et~al.},\ }\href@noop {} {\bibinfo {title} {Keras}},\
	\bibinfo {howpublished} {\url{https://keras.io}} (\bibinfo {year}
	{2015})\BibitemShut {NoStop}%
	\bibitem [{\citenamefont {Guillemin}\ and\ \citenamefont
		{Pollack}(2000)}]{Guillemin:2000ti}%
	\BibitemOpen
	\bibfield  {author} {\bibinfo {author} {\bibfnamefont {V.}~\bibnamefont
			{Guillemin}}\ and\ \bibinfo {author} {\bibfnamefont {A.}~\bibnamefont
			{Pollack}},\ }\href@noop {} {\emph {\bibinfo {title} {{Differential
					Topology}}}}\ (\bibinfo  {publisher} {AMS Chelsea Publishing},\ \bibinfo
	{address} {Providence, Rhode Island},\ \bibinfo {year} {2000})\BibitemShut
	{NoStop}%
	\bibitem [{\citenamefont {Edson}\ \emph {et~al.}(2019)\citenamefont {Edson},
		\citenamefont {Bunder}, \citenamefont {Mattner},\ and\ \citenamefont
		{Roberts}}]{Edson2019}%
	\BibitemOpen
	\bibfield  {author} {\bibinfo {author} {\bibfnamefont {R.~A.}\ \bibnamefont
			{Edson}}, \bibinfo {author} {\bibfnamefont {J.~E.}\ \bibnamefont {Bunder}},
		\bibinfo {author} {\bibfnamefont {T.~W.}\ \bibnamefont {Mattner}},\ and\
		\bibinfo {author} {\bibfnamefont {A.~J.}\ \bibnamefont {Roberts}},\ }\href
	{https://doi.org/10.1017/S1446181119000105} {\bibfield  {journal} {\bibinfo
			{journal} {ANZIAM Journal}\ }\textbf {\bibinfo {volume} {61}},\ \bibinfo
		{pages} {270} (\bibinfo {year} {2019})},\ \Eprint
	{https://arxiv.org/abs/1902.09651} {arXiv:1902.09651} \BibitemShut {NoStop}%
\end{thebibliography}

\end{document}